\documentclass{article}

\PassOptionsToPackage{numbers, compress}{natbib}
\PassOptionsToPackage{table,dvipsnames}{xcolor}

    \usepackage[final]{neurips_2025}

\usepackage[utf8]{inputenc} 
\usepackage[T1]{fontenc}    
\usepackage{hyperref}       
\usepackage{url}            
\usepackage{booktabs}       
\usepackage{amsfonts}       
\usepackage{nicefrac}       
\usepackage{microtype}      
\usepackage{natbib}
\usepackage{graphicx}
\usepackage{adjustbox}
\usepackage{amsmath}
\usepackage{wrapfig}
\usepackage{longtable}
\usepackage{comment}
\usepackage{xspace}
\usepackage[textsize=tiny]{todonotes}
\usepackage[most]{tcolorbox}
\usepackage{makecell}
\usepackage{collcell}
\usepackage{booktabs}
\usepackage{graphicx}
\usepackage{xspace}
\usepackage{multirow}
\usepackage{makecell}
\usepackage{caption}
\usepackage{enumitem} 
\usepackage{diagbox}
\usepackage[toc,page]{appendix}
\usepackage{titletoc}
\usepackage{subcaption}

\definecolor{cyan}{RGB}{222, 255, 255}
\definecolor{realcyan}{RGB}{0, 230, 230}
\definecolor{columbiablue}{rgb}{0.61, 0.87, 1.0}
\definecolor{nicegreen}{RGB}{77, 204, 77}
\definecolor{nicered}{RGB}{204, 77, 77}

\newcommand{\webshepherd}{\textsc{Web-Shepherd}\xspace}
\newcommand{\dataset}{\textsc{WebPRM Collection}\xspace}
\newcommand{\benchmark}{\textsc{WebRewardBench}\xspace}
\definecolor{lightredshade}{HTML}{dea9a9}
\definecolor{lightgreenshade}{HTML}{bce3bd}
\definecolor{lightblueshade}{HTML}{cacbe8}
\definecolor{MyDarkBlue}{rgb}{0,0.08,1}
\definecolor{MyDarkGreen}{rgb}{0.02,0.6,0.02}
\definecolor{MyDarkRed}{rgb}{0.8,0.02,0.02}
\definecolor{MyDarkOrange}{rgb}{0.40,0.2,0.02}
\definecolor{MyPurple}{RGB}{111,0,255}
\definecolor{MyRed}{rgb}{1.0,0.0,0.0}
\definecolor{MyGold}{rgb}{0.75,0.6,0.12}
\definecolor{MyDarkgray}{rgb}{0.66, 0.66, 0.66}

\definecolor{MyYellow}{rgb}{254, 246, 170}
\definecolor{MyBlue}{rgb}{170, 217, 251}
\definecolor{LuneBlue}{rgb}{0.11, 0.11, 0.43}

\newcommand{\eg}{e.g.,\xspace}
\newcommand{\ie}{i.e.,\xspace}

\newcommand{\method}{\textsc{Web-Shepherd}\xspace}
\newcommand{\wprmbench}{\textsc{WebRewardBench}}

\newcommand{\gpto}{GPT-4o\xspace}

\newcommand{\qwenvllarge}{Qwen-2.5-VL-72B\xspace}
\newcommand{\gptomini}{GPT-4o-mini\xspace}

\usepackage{amssymb}%
\usepackage{pifont}%
\newcommand{\greencheck}{\color{nicegreen}\ding{51}}
\newcommand{\redx}{\color{nicered}\ding{55}}

\hypersetup{
    colorlinks = true,
    linkbordercolor = {white},
    linkcolor = blue!60!black,
    citecolor = blue!60!black,
    urlcolor = blue!60!black
}

\usepackage{fontawesome5}

\title{\vspace{-0.3cm}\includegraphics[width=1.0cm]{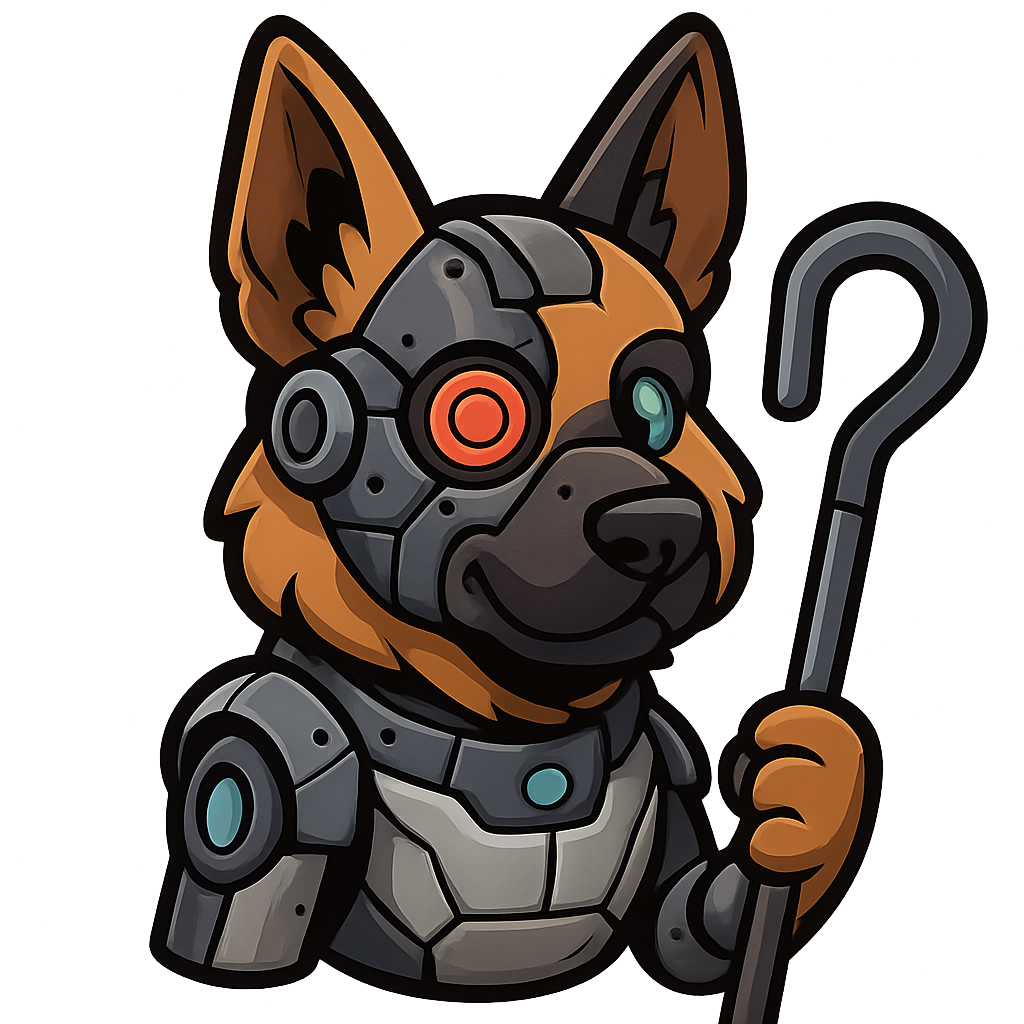}\webshepherd: \\
Advancing PRMs for Reinforcing Web Agents}

\author{
    Hyungjoo Chae\textsuperscript{1}\thanks{Equal contribution. This work was done while Hyungjoo Chae was at Yonsei University. } \quad
    Sunghwan Kim\textsuperscript{2}\footnotemark[1] \quad
    Junhee Cho\textsuperscript{2}\footnotemark[1] \quad \\
    \textbf{Seungone Kim}\textsuperscript{3} \quad 
    \textbf{Seungjun Moon}\textsuperscript{2} \quad 
    \textbf{Gyeom Hwangbo}\textsuperscript{2} \quad
    \textbf{Dongha Lim}\textsuperscript{2} \quad
    \textbf{Minjin Kim}\textsuperscript{2} \quad \\
    \textbf{Yeonjun Hwang}\textsuperscript{2} \quad 
    \textbf{Minju Gwak}\textsuperscript{2} \quad
    \textbf{Dongwook Choi}\textsuperscript{2} \quad
    \textbf{Minseok Kang}\textsuperscript{2} \quad
    \textbf{Gwanhoon Im}\textsuperscript{2} \quad \\
    \textbf{ByeongUng Cho}\textsuperscript{2} \quad
    \textbf{Hyojun Kim}\textsuperscript{2} \quad
    \textbf{Jun Hee Han}\textsuperscript{2} \quad 
    \textbf{Taeyoon Kwon}\textsuperscript{2} \quad \\
    \textbf{Minju Kim}\textsuperscript{2} \quad
    \textbf{Beong-woo Kwak}\textsuperscript{2} \quad 
    \textbf{Dongjin Kang}\textsuperscript{2} \quad
    \textbf{Jinyoung Yeo}\textsuperscript{2}\thanks{Corresponding author. Contact: \texttt{hchae36@gatech.edu} and \texttt{jinyeo@yonsei.ac.kr}.} \\
    \textsuperscript{1}Georgia Institute of Technology \\
    \textsuperscript{2}Department of Artificial Intelligence, Yonsei University \quad
    \textsuperscript{3}Carnegie Mellon University
}

\begin{document}

\maketitle

\begin{abstract}
Web navigation is a unique domain that can automate many repetitive real-life tasks and is challenging as it requires long-horizon sequential decision making beyond typical multimodal large language model (MLLM) tasks.
Yet, specialized reward models for web navigation that can be utilized during both training and test-time have been absent until now. Despite the importance of speed and cost-effectiveness, prior works have utilized MLLMs as reward models, which poses significant constraints for real-world deployment. To address this, in this work, we propose the first process reward model (PRM) called \method which could assess web navigation trajectories in a step-level. To achieve this, we first construct the \dataset, a large-scale dataset with 40K step-level preference pairs and annotated checklists spanning diverse domains and difficulty levels. Next, we also introduce the \benchmark, the first meta-evaluation benchmark for evaluating PRMs. In our experiments, we observe that our \method achieves about 30 points better accuracy compared to using GPT-4o on \benchmark. 
Furthermore, when testing on WebArena-lite by using GPT-4o-mini as the policy and \method as the verifier, we achieve 10.9 points better performance, in 10$\times$ less cost compared to using GPT-4o-mini as the verifier. 
Our model, dataset, and code are publicly available at \href{https://github.com/kyle8581/Web-Shepherd}{LINK}.
\end{abstract}

\begin{figure}[!h]
    \centering
    \includegraphics[width=0.98\linewidth]{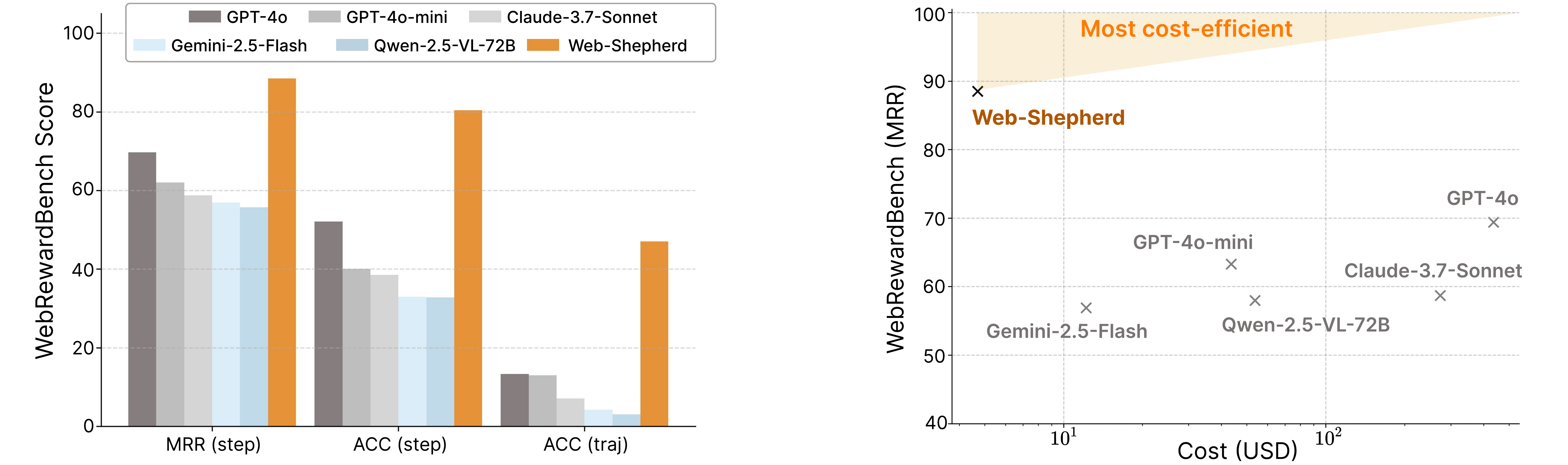}
    \caption{Performance and cost-efficiency of \method (3B). \method achieves the state-of-the-art performance while requiring significantly lower cost compared to existing baselines.}
    \label{fig:performance}
\end{figure}
\section{Introduction}

Web browsers serve as a common interface for countless digital tasks, making automation in this space a natural focus for recent advances in intelligent agents.
Recent advances in multimodal large language models (MLLMs) have enabled agents to handle basic web interactions, such as retrieve addresses from map services or navigating simple webpages~\citep{zhou2023webarena,pan2024webcanvas}.
 However, current agents remain highly unreliable, often exhibiting brittle behaviors such as repeatedly entering the same query when encountering minor issues, eventually failing the task~\citep{koh2024visualwebarena,wang2024agent, chae2025web}. This unreliability primarily stems from the long-horizon nature of web navigation, requiring agents to reason across multiple steps and maintain goal-directed planning, which MLLMs often find challenging~\citep{xue2025illusion}.
 Hence, to create a better performing web agent, there is a need for better learning methods and inference-time algorithms.

One effective method that allowed large language models (LLMs) to perform well across various tasks is using a reward model to perform search at test-time (\eg Best-of-$n$), or using it for Reinforcement Learning (\eg RLHF). However, specially trained reward models have been under-explored in the web navigation domain. Prior works such as \citet{pan2024autonomous} and \citet{koh2024tree} do not train separate reward models, but instead employ MLLMs as evaluators in inference-time algorithms, which has fundamental problems. First, using the evaluation from prompted MLLMs becomes a significant constraint in web navigation where speed and cost are crucially important. For example, using only GPT-4o for tree search on WebArena (consisting of 812 queries) requires approximately \$14,000, and running inference on one A100 takes 40 hours, which is a major obstacle to deploying MLLMs as web navigation agents in real-world scenarios. Additionally, throughout our experiments, we confirm that prompting MLLMs performs worse than trained reward models. In summary, considering speed, cost, and performance, specially designed reward models for web navigation are absolutely necessary.

To address these challenges, we present \method, which is, to the best of our knowledge, the \textbf{first reward model trained specifically for evaluating trajectories of web navigation}. In particular, \method is designed as a process reward model (PRM) rather than an outcome reward model (ORM), because unlike other domains, ORM cannot be integrated into test-time algorithms in web navigation. For example, in mathematics, an LLM can write multiple solutions and the ORM can choose one, but in web navigation, if an LLM makes eight attempts to book a plane ticket, the airplane ticket cannot be refunded, so decisions about which action to take must be made at the process level. Furthermore, even during training-time, PRM can provide more fine-grained reward signals, making it more reliable than ORM~\citep{lightman2023let,wang2024math}. \method employs a structured checklist that explicitly decomposes high-level user instructions into clear, interpretable subgoals. By referencing this checklist as evaluation criteria, \method accurately assesses step-level progress, enabling precise and robust guidance throughout agent trajectories.

The key contribution of this paper is that we also provide a \textbf{suite of training data and benchmark to test PRMs for web navigation}. First, we release the \dataset, which contains human-crafted instructions that covers diverse tasks across multiple difficulty levels. The notable feature of the \dataset is that it contains 40K step-level annotations for which action an agent should take and that each instruction contains an annotated checklist—structured sequences of subgoals that enable \method to make accurate judgments. Second, we release the \benchmark, the first meta-evaluation benchmark to assess PRMs in web navigation. The \benchmark allows practitioners to test newly proposed PRMs without running resource-intensive web navigation agents, enabling efficient testing of different design choices and conducting ablation experiments. \method achieves 85.0\% performance on the \benchmark (WebArena set), significantly outperforming GPT-4o-mini with prompting at 5.0\%. Furthermore, when using \method's reward as guidance in tree search on \gptomini policy, it achieves 34.55\% success rate on WebArena-lite~\citep{zhou2023webarena,liu2025visualagentbench}, confirming it is 10.9 points better in performance, and 10$\times$ more cost-effective than using GPT-4o-mini as the evaluator.

\section{Related Work}

\paragraph{MLLM-based web agents.}
Multimodal large language models (MLLMs) have emerged as powerful foundation models for web agents due to their strong generalization capabilities and adaptability to diverse interface. Previous work has leveraged MLLMs to complete web tasks via carefully designed instructions, often augmented with external tools (\eg grounding module or verification)~\citep{zheng2024seeact, zheng2023synapse, he2024webvoyagerbuildingendtoendweb, tao2023webwisewebinterfacecontrol} or workflow~\citep{wang2024agent}.
Moreover, other approaches have trained MLLM-based agents to imitate expert trajectories using next-token prediction objective~\citep{deng2024mind2web, cheng2024seeclickharnessingguigrounding, kil2024dualviewvisualcontextualizationweb}.
While these models perform well in-distribution, they often fail to generalize to unseen environments.
To overcome these challenges, recent research has increasingly focused on inference-time scaling~\citep{pan2024autonomous, koh2024treesearchlanguagemodel} or reinforcement learning (RL)~\citep{qi2025webrltrainingllmweb, bai2024digirltraininginthewilddevicecontrol, bai2025digiqlearningqvaluefunctions}, which enables agents to improve decision-making through reward feedback.

\paragraph{Inference-time scaling for web agents.}
Inference-time scaling has emerged as a crucial approach for multi-turn interactions in web environment.
Recent studies have explored techniques such as tree search~\citep{feng2023alphazero, koh2024treesearchlanguagemodel}, long chain-of-thought (CoT)~\citep{jaech2024openaio1, guo2025deepseek}, and incorporating verifiers or judges to enhance agent performance with natural language feedback~\citep{shinn2024reflexion, pan2024autonomousevaluationrefinementdigital}.
For example, \citet{pan2024autonomousevaluationrefinementdigital} use a prompting-based evaluator to assess whether a trajectory is successful; if not, they apply Reflexion~\citep{shinn2024reflexion} to retry based on the generated feedback.
Extending this direction, other work~\citep{koh2024tree, chae2025web} investigates an interesting direction that tries to search the optimal browsing path with a prompted value function and A*-like algorithm and world model.

\paragraph{Rewards for web navigation.}
Prior works rely on binary rewards (success or failure)~\citep{bai2024digirltraininginthewilddevicecontrol, bai2025digiqlearningqvaluefunctions} from rule-based evaluations that require human annotation and lacks scalability in dynamic web environments~\citep{zhou2023webarena, koh2024visualwebarena}.
To address these, recent studies have explored leveraging LLMs via prompting~\cite{he2024webvoyagerbuildingendtoendweb, pan2024autonomous} or training outcome reward models (ORMs)~\citep{qi2025webrltrainingllmweb}.
However, binary reward offers limited guidance for credit assignment, especially in long-horizon tasks.
To enable more informative feedback, the reasoning literature has introduced process reward models (PRMs), which assign step-level reward~\citep{lightman2023let, wang2024math}.
Building on this idea, recent work has explored using LLMs to estimate state-action values by prompting~\citep{koh2024treesearchlanguagemodel, putta2024agentqadvancedreasoning, dai2025advancingmobileguiagents}.
Nevertheless, the reliability and efficiency of MLLMs as process-level reward models remain underexplored.
In this work, we aim to develop a PRM for web agents to support effective learning and cost-efficient inference-time guidance.

\section{Preliminaries}
\begin{wrapfigure}{r}{0.35\textwidth}
    \centering
    \vspace{-1em}
    \includegraphics[width=0.32\textwidth]{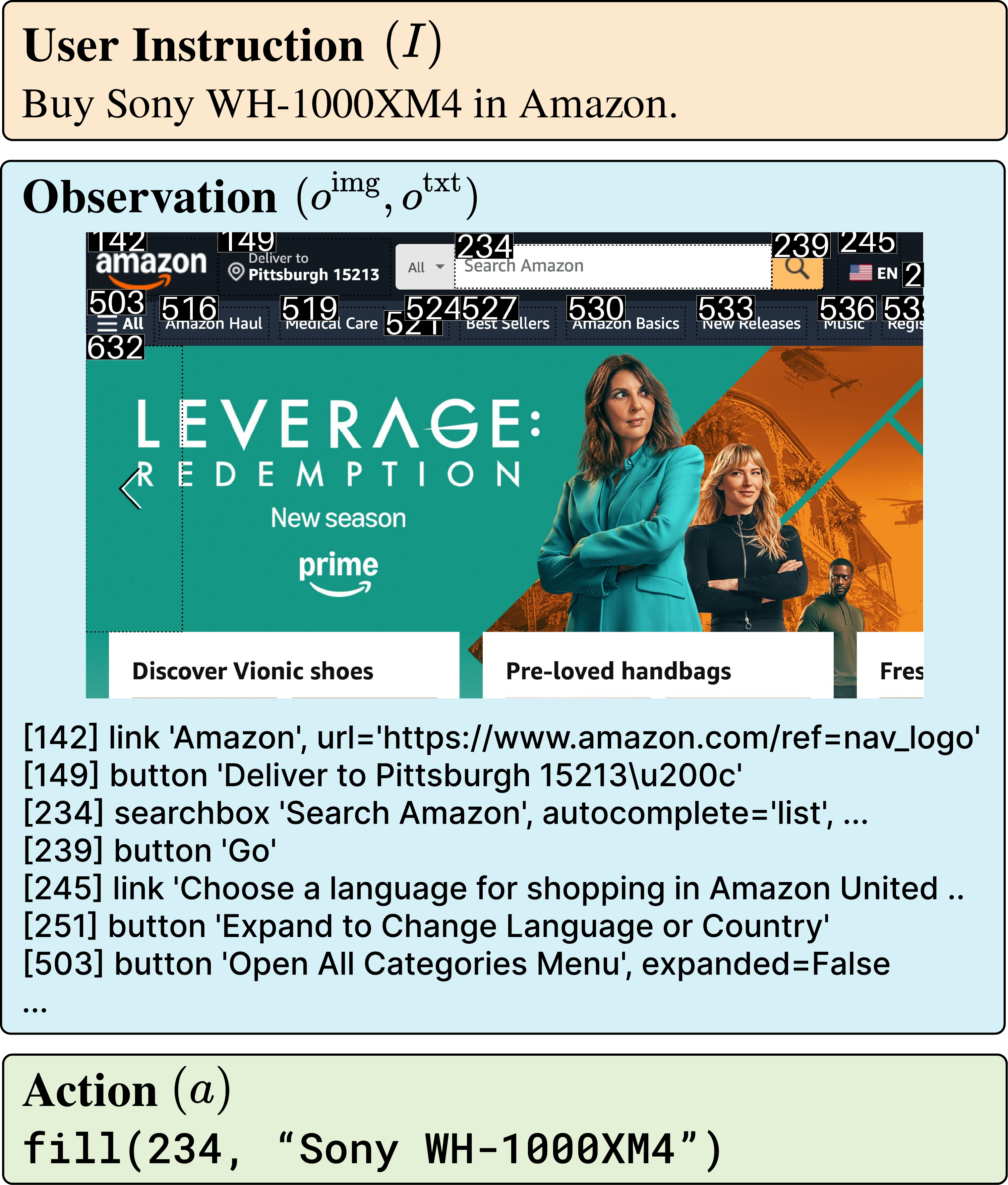}
    \caption{
        Example of web navigation under a POMDP.
        }
    \label{fig:task_example}
    \vspace{-1em}
\end{wrapfigure}
We formulate the web navigation problem as a partially observable Markov decision process (POMDP)
defined by the tuple $(\mathcal{S}, \mathcal{A}, \mathcal{O}, T, R)$, where $\mathcal{S}$ is the set of environment states, $\mathcal{A}$ is the set of agent actions, $\mathcal{O}$ is the set of observations, $T(s' \mid s, a)$ is the transition function, $R(s, a)$ is the reward function.
At each time step $t$, the agent receives a browser-rendered observation $o_t \in \mathcal{O}$ that only partially reflects the true underlying state $s_t \in \mathcal{S}$. In the context of web environments, $o_t$ consists of two modalities: (1) an accessibility tree  $o_t^{\text{txt}}$, a text sequence of intractable elements that captures the hierarchical and semantic structure of the webpage elements~\citep{zhou2023webarena, drouin2024workarena}, and (2) a rendered screenshot image $o_t^{\text{img}}$ depicting the visual appearance of the browser~\citep{koh2024visualwebarena}. Given these observations, the agent selects an action $a_t \in \mathcal{A}$ from a discrete set of browser-level commands, including operations such as \texttt{click}(i), \texttt{scroll}(d), and \texttt{type}(``text''), where $i$ is the index of a DOM or accessibility node, and $d$ denotes a scroll direction or offset. The agent's goal is to select actions that maximize the expected reward over a trajectory $\tau = (o_1, a_1, \dots, o_T)$.

\section{\dataset}
The major challenge of building a PRM in web navigation is the lack of a training dataset.
To address this, we collect \dataset, the first dataset for training PRMs for web agents. 
Our goal is to collect a dataset $\mathcal{D}$ that contains $(I, O, C, A^+, A^-)$, where $A^+$ is a sequence of chosen actions $(a^+_1, a^+_2, ..., a^+_n)$, \ie an expert trajectory, and $A^-$ is a sequence of rejected actions $(a^-_1, a^-_2, ..., a^-_n)$ along with the checklist $C$, observations $O = (o_1, o_2, ..., o_n)$, and user instruction $I$. 

\begin{figure}
    \centering
    \includegraphics[width=0.96\linewidth]{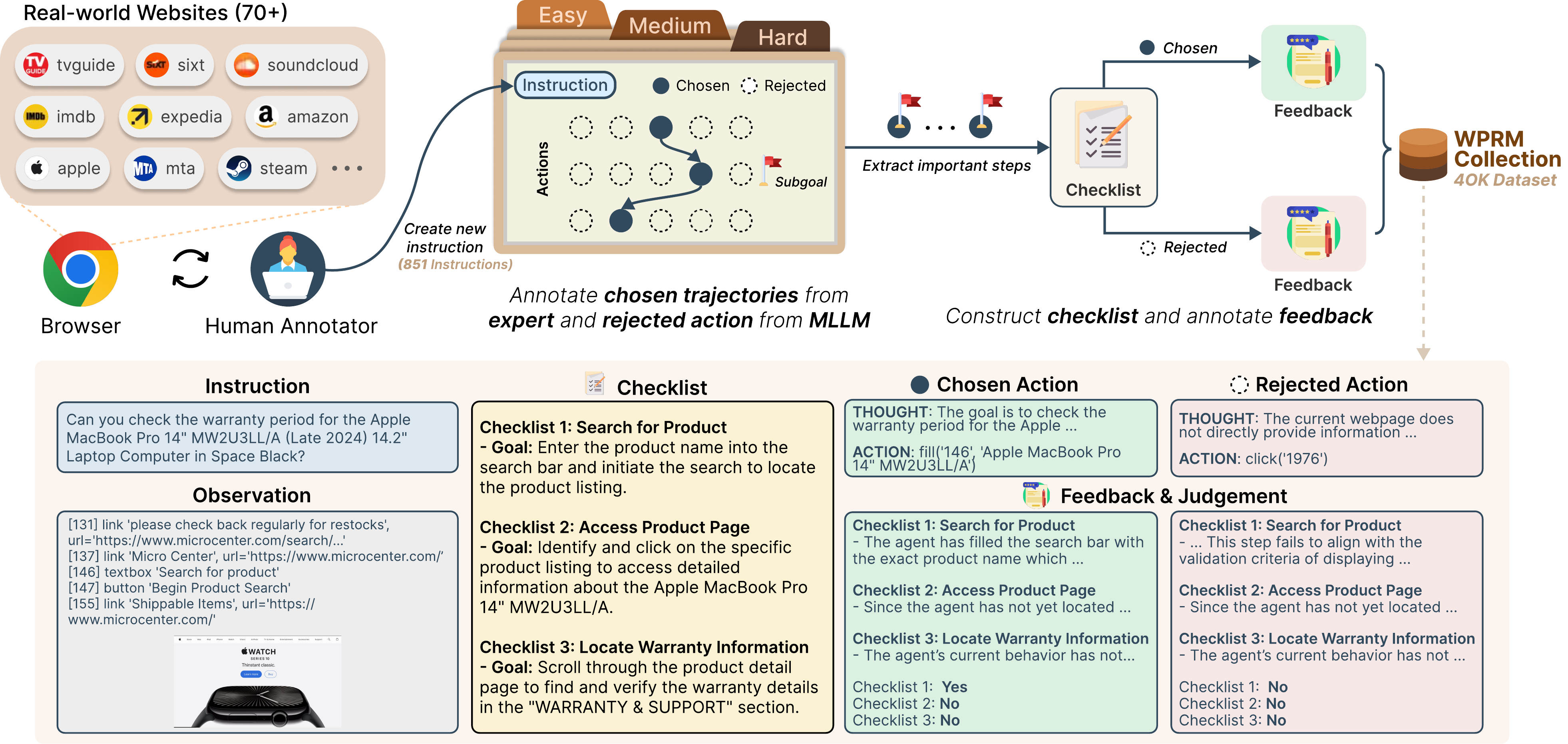}
    \caption{Overview of the dataset collection process of \dataset (top) and an example instance of our dataset (bottom).}
    \label{fig:main_dataset}
\end{figure}

\subsection{Collecting User Instruction and Expert Trajectory}

From human experts, we collect user instructions $I$ and the chosen actions $A^+$.
We select websites that permit access via \textit{playwright} from the pool of sites used in Mind2Web~\citep{deng2024mind2web}. Prior to annotation, all annotators participated in a three-hour training session designed to familiarize them with our annotation toolkit and to clarify the differences between human and agent browsing behaviors. Following annotation, all collected data were reviewed by a panel of 10 human evaluators to ensure quality and consistency.
During this process, we filtered out invalid trajectories that could not be reproduced, as well as vague instructions prone to misinterpretation. Annotators were instructed to craft instructions $I$ spanning three difficulty levels: easy, medium, and hard.

\subsection{Annotating Checklist and Rejected Action}
\paragraph{Checklist.}
To mitigate bias toward specific websites and reduce sensitivity to action orderings, we construct coarse-grained checklists that emphasize meaningful task progress over exact execution steps.
For example, fine-grained actions such as \textit{filter A} and \textit{filter B} are abstracted into a higher-level subgoal like \textit{filtering}.
This abstraction enables the model to generalize across semantically equivalent strategies.
Given an instruction $I$ and an expert trajectory $A^+$, we use GPT-4o to generate subgoal analysis and corresponding checklists.

\paragraph{Rejected actions.}
\begin{wrapfigure}{r}{0.48\textwidth}
    \centering
    \vspace{-1em}
    \includegraphics[width=0.48\textwidth]{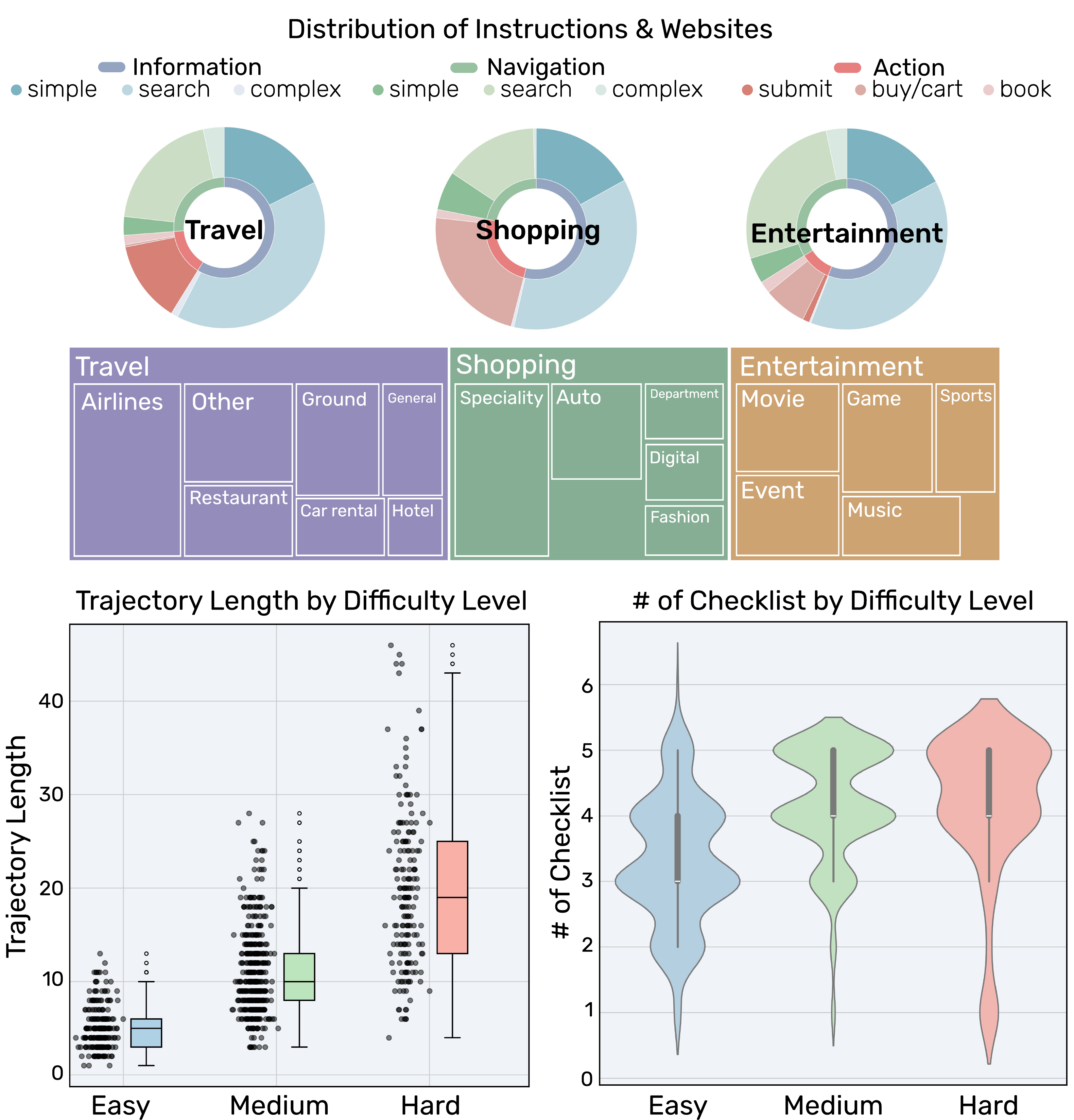}
    \caption{
        Statistics of \dataset.
        }
    \label{fig:dataset_stat}
    \vspace{-3em}
\end{wrapfigure}
To collect rejected actions $a_t^{-}$, we sample 5 candidate actions from diverse policies and select those that differ from the expert action $a_t^{+}$.
However, some of these alternatives may correspond to valid but different actions toward task completion (\eg \texttt{fill(423, ``Sony Camera'')} vs. \texttt{click(search\_box)}), rather than being truly suboptimal or incorrect.
To minimize such cases, we apply rule-based filtering and collect up to five rejected actions $a^{-}_t$ per expert action $a^{+}_t$.
More details about dataset construction are provided in Appendix~\ref{app:wprm_collection}.

\subsection{Dataset Statistics}
As shown in Figure~\ref{fig:dataset_stat}, we analyze two key aspects across difficulty levels: the length of agent trajectories and the number of checklist subgoals.
The left violin plot illustrates that trajectory length increases with difficulty. Easy tasks generally require fewer steps (median $\approx$ 5), whereas medium tasks show more variability (median $\approx$ 9), and hard tasks involve significantly longer trajectories (median $\approx$ 20), with some exceeding 40 steps. This indicates that our difficulty annotation effectively reflects the complexity and required interaction depth.
The right violin plot shows that the number of checklist items also grows with task difficulty, though the range is more concentrated. Easy tasks typically involve 3–4 checklist items, while medium and hard tasks consistently require 4–5 subgoals. 

\begin{figure*}[t!]
\centering
\includegraphics[width=0.95\linewidth]{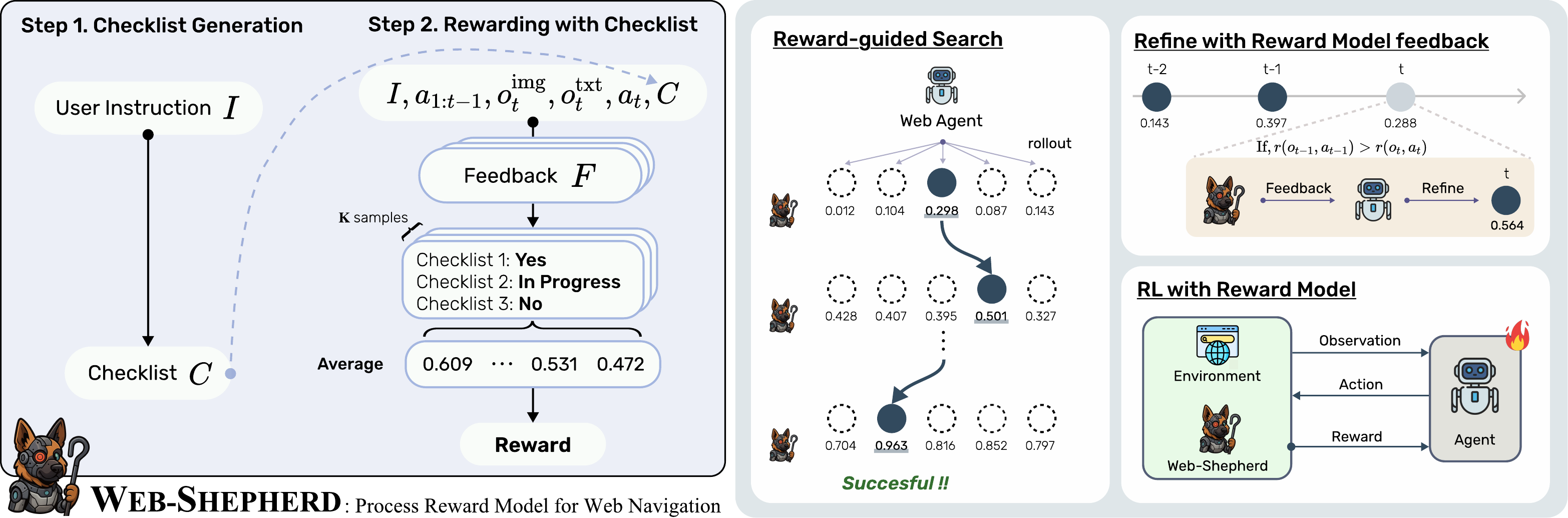}
\caption{Overview of \webshepherd (left) and its diverse use cases (right).}
\label{fig:web_shepherd}
\end{figure*}

\section{\webshepherd} 
In this section, we introduce \webshepherd, a process reward model designed to provide dense and reliable supervision to web agents and enable more informative credit assignment.
We train \webshepherd on the \dataset to support two key functionalities: (1) generating task-specific checklists, and (2) assigning rewards based on checklist completion.

\subsection{Step 1: Checklist Generation}
As illustrated in Figure~\ref{fig:web_shepherd}, \webshepherd first generates a checklist that outlines key intermediate milestones for achieving the user's goal. Given an instruction $I$, it produces a checklist $C$ comprising a sequence of natural language subgoals $(g_1, g_2, \cdots, g_k)$. This checklist then serves as the foundation for reward prediction, enabling \method to track progress toward the goal.
We further investigate the impact of checklist quality in Section~\ref{ssec:checklist_quality}.

\subsection{Step 2: Reward Modeling with Checklist}

\paragraph{Reward modeling as next-token prediction.}
To leverage the internal reasoning capabilities of MLLMs, we choose next-token prediction as our learning objective~\citep{mahan2024generative}. We optimize the language modeling loss over targets formed by concatenating the feedback $F$ and the judgment $J$, treating the full sequence as a coherent response. For example, given an input consisting of a checklist $C$, an observation $o$, and an answer $a$, the model is trained to generate the corresponding feedback and judgment in an auto-regressive manner. The loss is defined as:

\begin{equation}
\mathcal{L}_{\text{NTP}} = - \sum_{t} \log P_{\theta}(y_t \mid y_{<t}, C, o, a),
\end{equation}

where $y = [F; J]$ denotes the concatenated feedback and judgment tokens. This objective encourages the model to learn to evaluate the trajectories based on the checklist with reasoning and provide valuable feedback that explains the evaluation.

\paragraph{Scoring process reward.}
Since the reward is predicted via token generation, the output resides in a discrete space.
To obtain a continuous reward signal, several mapping strategies can be employed. One approach is to sample multiple output sequences and compute the average reward. 
Alternatively, we employ a verbalizer~\citep{hu2022verbalizer} to estimate soft probabilities over label tokens (\eg ``Yes'', ``No'', and ``In Progress'') using the logits from the LM head.
At inference time, \webshepherd generates the feedback $F\sim P(\cdot|I,C,o,a)$ and compute the reward for each checklist item using the probabilities of ``Yes'' and ``In Progress'' tokens follow:
\begin{equation}
r_k(o,a) = \frac{1}{L}\sum^{L}_{l}{P(\text{``Yes''}|I,C,o,a,F) + 0.5\times P(\text{``In Progress''}|I,C,o,a,F)},
\end{equation}
where $L$ denotes the number of checklist and $r_k$ is the score assigned to the  $k^{\text{th}}$ response. The final reward is computed as the average: $r(o,a)=\sum_{k=1}^{K}r_k(o,a)$.
We provide an empirical comparison of different scoring strategies in Appendix~\ref{app:scoring_strategy}.

\section{Experiments}
To evaluate the effectiveness of PRMs for web navigation, we conduct comprehensive experiments in assigning process-level reward for web agents, focusing on both the accuracy of reward assignment and the utility of those rewards in improving agent performance.

\subsection{\wprmbench}
\label{ssec:main_result}
In developing PRMs, a reliable benchmark (\eg RewardBench~\citep{lambert2024rewardbench}) is essential for evaluating their performance. However, there does not yet exist a benchmark specifically designed to evaluate how accurately models assign process rewards to web agents' trajectories.
To address this, we introduce \wprmbench, a benchmark that directly measures the accuracy of predicted rewards. 

\subsubsection{Setup}
\paragraph{Benchmark construction.}
We use two data sources, Mind2Web and WebArena, to obtain user instructions for web navigation tasks.
For Mind2Web, we utilize the expert demonstrations provided in the dataset. In contrast, since expert trajectories are unavailable in WebArena, we manually annotate them. As a result, we obtain 69 instances from WebArena and 707 instances from Mind2Web.
To construct a reliable benchmark for evaluating PRMs, we follow the setup of \citet{kim2024evaluating} and collect preference pairs $(o_t, a^+_t, \{a^{-}_{(t,i)}\}_{i=1}^4\})$, where each observation $o_t$ is paired with one chosen action and four rejected actions. Additionally, we provide reference checklists for each tasks to ensure fair and consistent evaluation.
Further details on benchmark construction are provided in Appendix~\ref{app:webprmbench_detail}.

\paragraph{Metrics.}
We evaluate process reward prediction using the following three metrics:
(1) Mean Reciprocal Rank (MRR): The average of the reciprocal ranks of the preferred action in the list of all candidate actions sorted by predicted reward. A higher MRR indicates that the model consistently ranks the preferred action closer to the top.
(2) Step Accuracy (Acc. step): The proportion of steps where the model assigns the highest predicted reward to the preferred action $a^+_t$ among the five candidates.
(3) Trajectory Accuracy (Acc. traj): The proportion of full trajectories where the model ranks $a^+$ highest at every step among the candidate actions.

\paragraph{Baselines.}
Prior work has leveraged prompted MLLMs to obtain process-level rewards by exploiting their reasoning and image understanding capabilities~\citep{chae2025web, koh2024tree}. 
Following this approach, we construct baselines using representative MLLMs from both open-source and closed-source categories. For open-source models, we use \gptomini and \gpto; for closed-source models, we adopt \qwenvllarge, which are widely used in recent literature.

\paragraph{Implementation of \method.}
We train \method on our dataset using the following base models: for text-only settings, we use Qwen2.5-3B~\citep{qwen2.5} and Qwen3-8B~\citep{yang2025qwen3technicalreport}; for multimodal settings, we use Qwen2.5-VL-3B~\citep{bai2025qwen2}. All models are trained for 3 epochs using LoRA~\citep{hu2022lora}.

\begin{table*}
\centering
\caption{Evaluation results on \wprmbench. \textbf{T}: text observation, \textbf{I}: image observation.}

\resizebox{0.98\textwidth}{!}{
\begin{tabular}{lcccccccccc}
\toprule
\multirow{3}{*}{\textbf{Model}} & \multirow{3}{*}{\textbf{Inputs}} & \multirow{3}{*}{\textbf{Checklist}} & \multicolumn{6}{c}{\textbf{Mind2Web}} & \multicolumn{2}{c}{\textbf{WebArena}}
\\
\cmidrule(lr){4-9}\cmidrule(lr){10-11}  
&  &  & \multicolumn{2}{c}{\textit{Cross-Task}} & \multicolumn{2}{c}{\textit{Cross-Website}} & \multicolumn{2}{c}{\textit{Cross-Domain}} & \multicolumn{2}{c}{\textit{Test}} \\ 
& & & MRR & $\text{Acc. (traj)}$ & MRR & Acc. (traj) & MRR & Acc. (traj) & MRR & Acc. (traj)\\
\midrule
\multirow{4}{*}{\gptomini} & \textbf{T} & \redx & 47.5 & 0.0 & 47.6 & 13.5 & 45.4 & 0.8 & 34.4 & 5.0 \\
& \textbf{T} & \greencheck & 63.9 & 5.0 & 66.1 & 12.8 & 63.3 & 12.4 & 60.0 & 15.0 \\
& \textbf{T} + \textbf{I} & \redx & 48.2 & 0.0 & 49.3 & 0.0 & 49.5 & 0.8 & 38.7 & 0.0 \\
& \textbf{T} + \textbf{I} & \greencheck & 58.8 & 2.5 & 64.4 & 5.1 & 63.0 & 4.1 & 53.8 & 5.0 \\
\midrule
\multirow{4}{*}{\gpto} & \textbf{T} & \redx & 56.9 & 5.0 & 55.8 & 2.6 & 59.8 & 3.3 & 59.2 & 15.0 \\
& \textbf{T} & \greencheck & 67.4 & 7.5 & 70.3 & 5.1 & 70.2 & 11.6 & 69.7 & 15.0 \\
& \textbf{T} + \textbf{I} & \redx & 52.5 & 5.0 & 52.2 & 0.0 & 52.8 & 1.7 & 49.7 & 5.0 \\
& \textbf{T} + \textbf{I} & \greencheck & 62.4 & 5.0 & 68.1 & 15.4 & 65.1 & 6.6 & 59.7 & 10.0 \\

\midrule
\multirow{4}{*}{\qwenvllarge} & \textbf{T} & \redx & 55.7 & 5.0 & 51.8 & 0.0 & 54.2 & 1.7 & 54.6 & 5.0 \\
& \textbf{T} & \greencheck & 59.4 & 0.0 & 62.4 & 0.0 & 57.9 & 1.7 & 52.3 & 5.0 \\
& \textbf{T} + \textbf{I} & \redx & 50.1 & 2.5 & 47.6 & 0.0 & 49.8 & 0.8 & 43.1 & 0.0 \\
& \textbf{T} + \textbf{I} & \greencheck & 52.9 & 2.5 & 53.5 & 2.6 & 52.0 & 2.5 & 47.3 & 0.0\\
 \midrule
 \midrule

\multirow{2}{*}{\webshepherd (3B)} &  \textbf{T} & \greencheck & 87.6 & 55.0 & \textbf{88.0} & 43.6 & 87.2 & 47.1 & 91.1 & 60.0 \\
 & \textbf{T+I} & \greencheck &  85.0 & 42.5 & 87.3 & 41.0 &  84.4 & 37.2 & 92.5 & 65.0 \\
 \midrule
\method (8B) & \textbf{T} & \greencheck & \textbf{88.3} & \textbf{57.5} & 87.9 & \textbf{51.3} &  \textbf{91.3} & \textbf{61.2} & \textbf{97.8} & \textbf{85.0} \\

\bottomrule
\end{tabular}
}
\label{tab:main_webprmbench}
\end{table*}

\subsubsection{Results}
\paragraph{MLLMs struggle with assigning correct process rewards.}
We evaluate the ability of models to accurately assign process rewards on \benchmark under different input types (text only vs. text and image) and with or without using the checklist.
As shown in Table~\ref{tab:main_webprmbench}, state-of-the-art MLLMs struggle to provide reliable rewards for web navigation tasks.\footnote{Step accuracy is omitted due to the limited space. We provide the full results in Appendix~\ref{app:additional_result}.} 
This limitation is particularly evident in the trajectory accuracy metric. In this measure, models frequently fail to assign correct rewards consistently at each time step within a single task.
In contrast, \method significantly outperforms all baselines, demonstrating a substantial performance gap across all benchmark settings.

\begin{wrapfigure}{r}{0.4\textwidth}
    \centering
    \vspace{-0.5em}
    \includegraphics[width=0.39\textwidth]{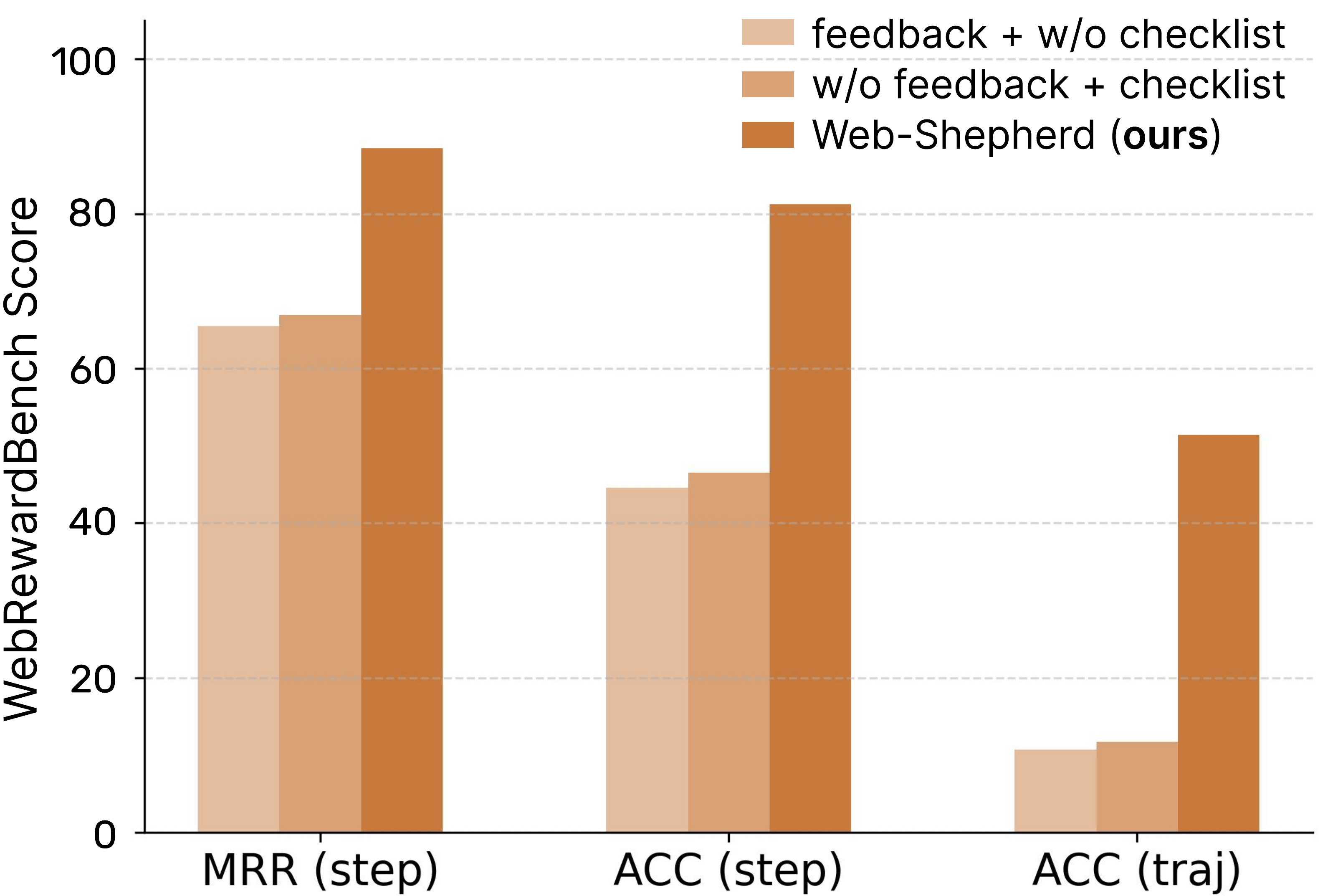}
    \caption{
        Results of ablation study with \method (3B).
        }
    \label{fig:ablation}
    \vspace{-2em}
\end{wrapfigure}

\paragraph{Checklist allows reliable reward assignment.}
Table~\ref{tab:main_webprmbench} demonstrates that both baseline and our models benefit significantly from the checklist in assigning rewards.
Checklists lead to more accurate and consistent reward assignments, as evidenced by improvements in trajectory accuracy across all baselines.
These results suggests that checklists serve as valuable guidance, helping models maintain coherence in predicting the process reward.
Furthermore, as shown in Figure~\ref{fig:ablation}, when we conduct ablation studies with models that are trained to 
either assign rewards without checklists or use checklists without feedback,
     we observe a substantial performance drop. These findings underscore the importance of both checklists and feedback for assigning reliable rewards.

\paragraph{Multimodal input does not always improve performance.}
Contrary to our expectations, incorporating multimodal input does not always lead to performance gains; in some cases, using multimodal input even degrades the performance. For example, when using \gpto as the reward model, we observe a notable improvement in trajectory accuracy only on the cross-website of Mind2Web subset. This observation is consistent with the findings of \citet{xue2025illusion}, which suggest that processing inputs from multiple modalities can introduce ambiguity and act as a source of noise, ultimately hindering the model performance.

\subsection{Reward-Guided Trajectory Search}
\label{ssec:BoN}

Reward-guided search using Best-of-$n$ (BoN) sampling offers a practical proxy for evaluating the capability of a reward model to guide policies~\citep{wang2024math, liu2025pairwise, liu2025inference}. Notably, it allows us to assess the potential for reward overoptimization without relying on reinforcement learning.
In addition, it provides an effective approach to adapting an MLLM policy without fine-tuning~\citep{koh2024tree, chae2025web, gu2024webdreamer}.

\paragraph{Setup.}
We evaluate our approach on WebArena-lite and WorkArena~\citep{drouin2024workarena} in an online setting. WebArena-lite~\citep{liu2025visualagentbench} is a subset of WebArena~\citep{zhou2023webarena}, comprising 165 instructions with error-corrected judge code from the earlier version. WorArena is a remote-hosted benchmark
of 33 tasks based on the widely-used ServiceNow
platform. Among 5 action candidates sampled from the policy, the action that is assigned the highest reward is executed.
For the policy, we use \gptomini, and compare performance when guided by our proposed PRM versus a prompt-based PRMs. We report the success rate (SR), which measures the proportion of tasks in which the final state satisfies the condition.

\begin{table*}
\centering
\caption{Success rates of trajectory search with \gptomini{} and \gpto{} as policy on WebArena-lite.}
\resizebox{0.90\textwidth}{!}{
\begin{tabular}{llc|ccccc|c|c}
\toprule
\textbf{Policy} & \textbf{PRM} & \textbf{Checklist} & \textbf{Shopping} & \textbf{CMS} & \textbf{Reddit} & \textbf{GitLab} & \textbf{Map} & \textbf{Total} & \textbf{$\Delta$} \\ 
\midrule
\multirow{5}{*}{\gptomini} 
  & w/o Trajectory Search & N/A & 21.74 & 22.86 & 19.05 & 34.38 & 19.35 &  23.64 & -- \\
  \cmidrule(lr){2-10}
  & \multirow{2}{*}{GPT-4o-mini} 
    & \redx & 13.04 & 14.29 & 9.52 & 25.00 & 16.13 & 15.75 & $-7.89$ \\
  &  & \greencheck & 21.74 & 31.43 & 14.29 & 34.38 & 16.13 & 24.24 & $+0.60$ \\
  & \webshepherd (3B) & \greencheck & \textbf{32.61} & 37.14& 19.05 & 34.38 & 32.26& 32.12& $+8.48$\\
  & \webshepherd (8B) & \greencheck & 26.09 & \textbf{45.71} & \textbf{23.81} & \textbf{40.62} & \textbf{35.48} & \textbf{34.55} &  $\mathbf{+10.90}$ \\

\midrule
\multirow{4}{*}{\gpto} 
  & w/o Trajectory Search & N/A & 23.91 & 31.43 & 28.57 & \textbf{56.25}& 19.35 & 31.52 & --\\
  \cmidrule(lr){2-10}
  & GPT-4o-mini & \greencheck & 21.74 & 31.43 & 28.57 & 40.62 & 12.90  & 26.67 & $-4.85$\\
  & \webshepherd (3B) & \greencheck & 28.26& 37.14 & \textbf{47.62}& 53.12& 25.81& 36.97 & $+5.45$\\
  & \webshepherd (8B) & \greencheck & \textbf{30.43} & \textbf{42.86} & \textbf{47.62} & 46.88& \textbf{35.48} & \textbf{39.39} & $\mathbf{+7.87}$ \\
\bottomrule
\end{tabular}
}
\label{tab:main_inference_time}
\end{table*}

\paragraph{Main results.}
We present the results in Table~\ref{tab:main_inference_time}. Interestingly, when using \gptomini as the reward model, we observe a slight improvement in the \gptomini policy. However, overall performance degrades when \gpto is used as the policy model, dropping from 31.52 to 26.67. In contrast, applying \method leads to substantial performance gains for both the \gptomini and \gpto policies across nearly all domains. Notably, \method boosts the \gptomini's browsing performance from 23.64 to 34.55, which is about 3 points higher than \gpto without trajectory search. These results suggest that \method remains effective in the online setting, even when paired with a stronger policy model.

\begin{table}[t]
\centering
\caption{Success rates of trajectory search with GPT-4o-mini as policy on {WorkArena}.}
\resizebox{\linewidth}{!}{
\begin{tabular}{lc|cccccccc|c}
\toprule
\textbf{PRM} & \textbf{Checklist} & \textbf{Dashboard} & \textbf{Form} & \textbf{Knowledge} & \textbf{List-filter} & \textbf{List-sort} & \textbf{Menu} & \textbf{Service Catalog} & \textbf{Total} & \textbf{$\Delta$} \\
\midrule
w/o Trajectory Search & N/A & 50.00 & 0.00 & 10.00 & 0.00 & 5.00 & 25.00 & 2.22 & 9.39 & – \\
\midrule
GPT-4o-mini & \greencheck & 55.00 & {10.00} & 10.00 & 0.00 & 6.67 & 20.00 & 5.56 & 12.42 & $+3.03$ \\
Web-Shepherd (3B) & \greencheck & 57.50 & \textbf{14.00} & 10.00 & 0.00 & \textbf{10.00} & 10.00 & \textbf{11.11} & 14.85 & $+5.46$ \\
Web-Shepherd (8B) & \greencheck & \textbf{65.00} & \textbf{14.00} & \textbf{20.00} & \textbf{0.00} & \textbf{10.00} & \textbf{20.00} & {7.78} & \textbf{15.76} & $\mathbf{+6.37}$ \\
\bottomrule
\end{tabular}
}
\label{tab:workarena_trajectory_search}
\end{table}

\paragraph{Results on WorkArena.}
To assess the robustness across domains, we also evaluate our models on WorkArena, a benchmark completely out-of-domain for \textsc{Web-Shepherd}.
As shown in Table~\ref{tab:workarena_trajectory_search}, trajectory search guided by the PRM improves the success rate in WorkArena, 
where the Total score increases from 9.39 to 12.42 when comparing the baseline without trajectory search.
Moreover, our model consistently outperforms GPT-4o-mini across all domains except for \textit{Menu}.
We attribute the relatively low performance in the Menu domain to the complexity of its multi-level dropdowns and embedded search boxes, 
which cause the policy model to produce unreliable action candidates.

\paragraph{Can \method provide useful feedback?}
\begin{wraptable}[9]{r}{0.4\textwidth}
\centering
\vspace{-1.0em}
\caption{Results of refinement with feedback from \method using GPT-4o-mini as the policy on WebArena-lite.}
\resizebox{0.38\textwidth}{!}{
    \begin{tabular}{l|c|c}
    \toprule
      \textbf{Models}  & \textbf{SR} & $\Delta$ \\
      \midrule
       w/o refine &  23.64 &  --  \\
       \midrule
       \method (3B) & 26.67 &  $+3.03$  \\
       \method (8B) & 27.88 &  $+4.24$ \\
      \bottomrule
    \end{tabular}
    }
    \label{tab:reflexion}
\vspace{-0em}
\end{wraptable} 
To evaluate the effectiveness of the feedback generated by \method, we conduct experiments in which the agent performs \textit{step-wise refinement} using our feedback, similar to the Self-Refine~\citep{madaan2024self}.
Specifically, the agent refine current action with the feedback when its current reward is lower than the previous reward assigned by \method. 
Interestingly, contrary to previous findings by~\citet{chae2025web} suggesting that step-wise feedback from models is not helpful and may even be detrimental, we observe notable improvements when incorporating model feedback during refinement. A possible explanation is that \method not only learns the impact of actions but also identifies patterns that characterize suboptimal behavior.

\section{Discussion}
\subsection{The Impact of Checklist Quality in Reward Prediction}
\label{ssec:checklist_quality}
We assess the quality of checklists generated by both baseline models and \method using G-Eval~\citep{liu2023g_eval}, with GPT-4o as the evaluator. To ensure a reliable evaluation, we provide the reference checklist to the evaluator alongside each generated checklist. The details of G-Eval are provided in Appendix~\ref{app:checklist_quality}. As shown in Figure~\ref{fig:checklist_reward_relation} (left), all models, except \method (3B), generate high quality checklists. Notably, our model, which is trained solely for checklist generation, achieves the highest score. Motivated by this result, we also release a standalone version of the checklist generation model. 
To better understand the role of checklist quality, we analyze reward prediction performance using checklists from various sources: an early version of our model (A), our final models (B and C) and ground-truth checklists (D). In Figure~\ref{fig:checklist_reward_relation} (right), we observe that high-quality checklists lead to more reliable reward assignments. However, the results also suggest that model's capability imposes a natural ceiling on reward prediction performance, regardless of checklist quality.

\begin{figure}
    \centering
    \includegraphics[width=0.98\linewidth]{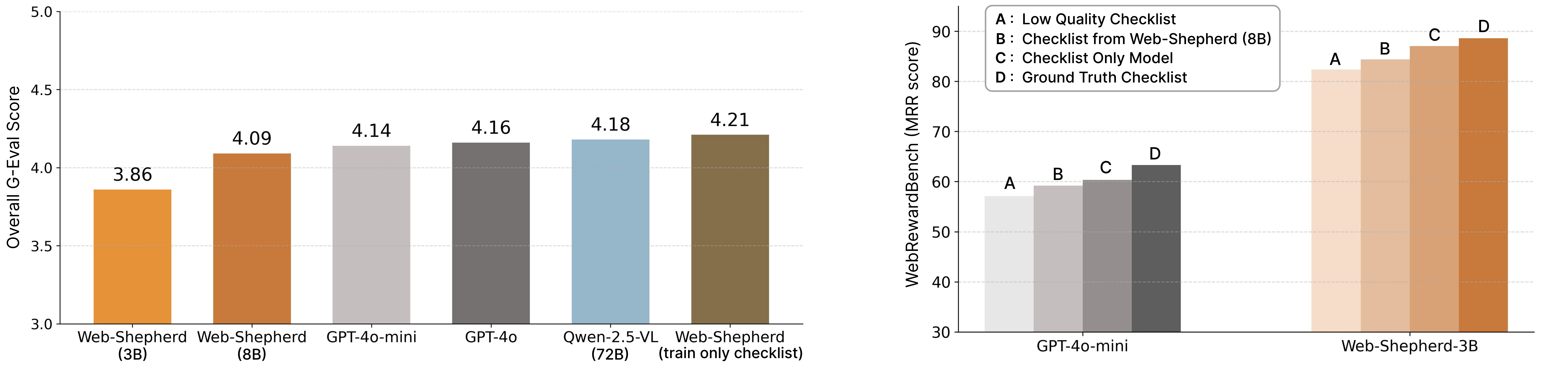}
    \caption{Evaluation of checklist quality (left) and its relationship to reward accuracy (right).
    }
    \label{fig:checklist_reward_relation}
\end{figure}

\subsection{Training Objective: Bradley-Terry Modeling vs. Generative Reward Modeling}
\begin{wrapfigure}{r}{0.44\textwidth}
    \centering
    \includegraphics[width=0.44\textwidth]{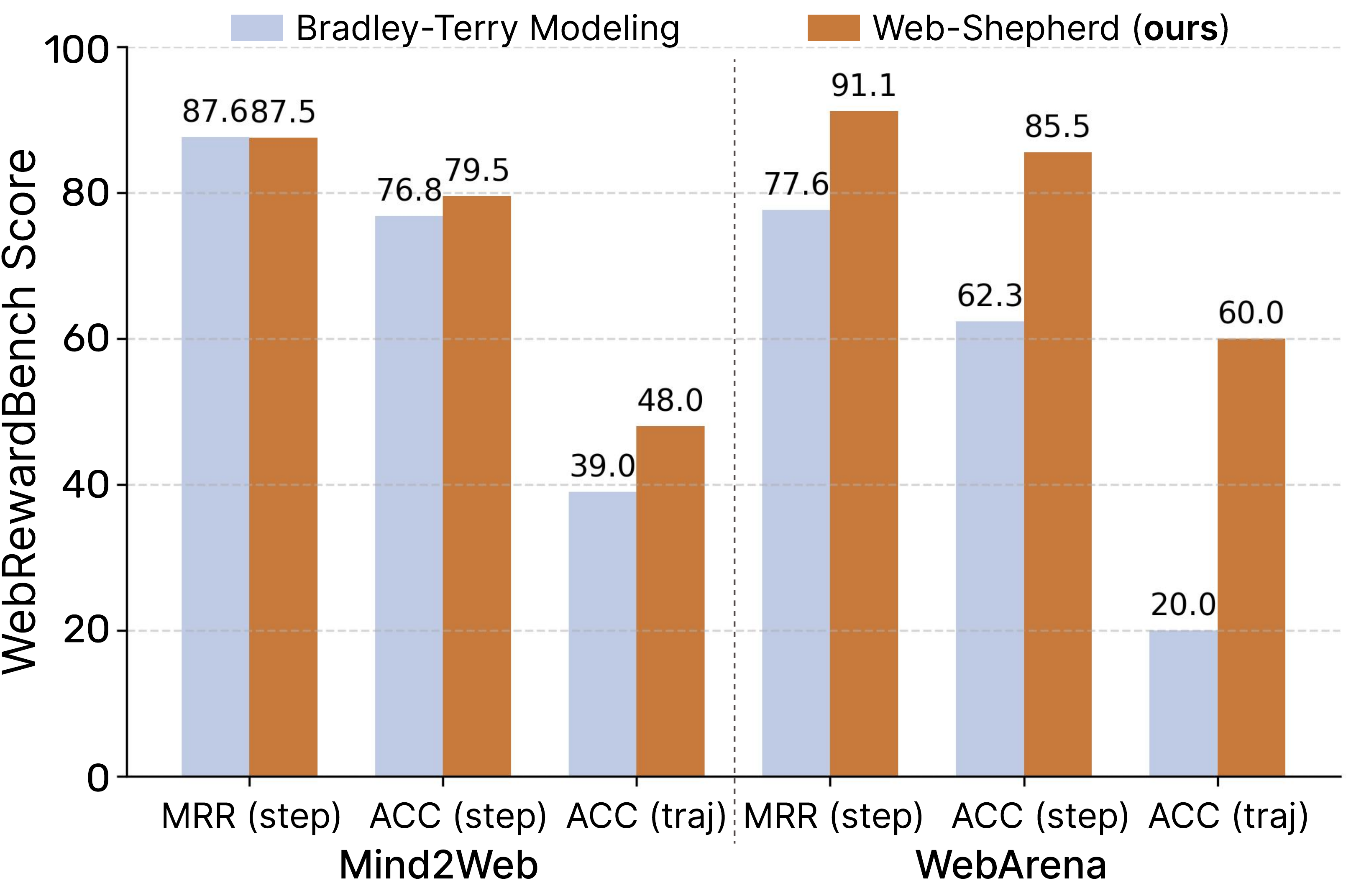}
    \caption{
        Analysis on the training objective.
        }
    \label{fig:train_objective}
    \vspace{-3em}
\end{wrapfigure}

The Bradley-Terry (BT) loss has been widely adopted as a training objective for learning reward models based on human preferences~\citep{ouyang2022training}.
However, its suitability for building PRMs in web navigation remains an open question.
To investigate this, we compare \method (3B) with a variant trained using the BT loss, with the identical training data.
As shown in Figure~\ref{fig:train_objective}, the BT-based model underperforms than ours, particularly in WebArena subset (out-of-distribution).
We find that the BT loss fails to effectively leverage the checklist for reward assignment, resulting in weaker sensitivity to task progress.
These findings suggest that BT modeling's key limitation\textemdash poor generalization observed across domains\textemdash also manifests in PRMs for web navigation.

\subsection{Cost Efficiency of \method}
We assess the cost efficiency of \method by comparing it to API-based models. For \method, costs are estimated using the hourly rate of an A100 80GB GPU instance (\$1.19/hour), combined with throughput measured via vLLM~\citep{kwon2023efficient}. Each instance averages 81,287 input and 1,953 output tokens and we compute cost of API-based models using publicly available prices. As shown in Figure~\ref{fig:performance} (right), \method delivers the best performance at the lowest cost per 1,000 instances—roughly 10$\times$ cheaper than GPT-4o-mini and 100$\times$ cheaper than GPT-4o.

\subsection{Data Scaling Law for PRM Training}
\begin{table*}[t]
\centering
\caption{Effect of instruction proportion and the number of rejected actions on model performance.}
\begin{subtable}[t]{0.49\textwidth}
\centering
\caption{Effect of instruction proportion.}
\resizebox{\linewidth}{!}{
\begin{tabular}{lcccc}
\toprule
\multirow{2}{*}{\makecell{\textbf{Proportion}\\\textbf{of instruction}}} &
\multicolumn{2}{c}{\textbf{Mind2Web}} &
\multicolumn{2}{c}{\textbf{WebArena}} \\
\cmidrule(lr){2-3} \cmidrule(lr){4-5}
 & MRR (step) & ACC (traj) & MRR (step) & ACC (traj) \\
\midrule
0.25 & 68.38 & 11.34 & 65.97 & 10.00 \\
0.5  & 77.46 & 21.54 & 73.57 & 15.00 \\
0.75 & 83.64 & 33.63 & 88.09 & 55.00 \\
Ours (3B) & \textbf{87.62} & \textbf{48.57} & \textbf{91.06} & \textbf{60.00} \\
\bottomrule
\end{tabular}
}
\label{tab:instruction_proportion}
\end{subtable}
\hfill
\begin{subtable}[t]{0.49\textwidth}
\centering
\caption{Ablation on the number of rejected actions.}
\resizebox{\linewidth}{!}{
\begin{tabular}{lcccc}
\toprule
\multirow{2}{*}{\makecell{\textbf{\# max}\\\textbf{ rejected actions}}} &
\multicolumn{2}{c}{\textbf{Mind2Web}} &
\multicolumn{2}{c}{\textbf{WebArena}} \\
\cmidrule(lr){2-3} \cmidrule(lr){4-5}
 & MRR (step) & ACC (traj) & MRR (step) & ACC (traj) \\
\midrule
1 & 71.42 & 12.79 & 63.04 & 10.00 \\
2 & 77.66 & 17.55 & 76.47 & 20.00 \\
3 & 79.70 & 24.91 & 77.46 & 20.00 \\
4 (Ours, 3B) & \textbf{87.62} & \textbf{48.57} & \textbf{91.06} & \textbf{60.00} \\
\bottomrule
\end{tabular}
}
\label{tab:rejected_action_ablation}
\end{subtable}
\end{table*}

We conduct analysis on the effect of the (1) number of instructions, and (2) number of rejected actions in the dataset on the performance of the PRM. Specifically, we construct datasets using the subset of WebPRMCollection 0.25, 0.5, and 0.75 percent of instruction and its corresponding chosen-rejected pairs and 1,2, and 3 number of max rejected actions. We trained variants of Web-Shepherd with these datasets using the same model (\ie Qwen-2.5-3B-Instruct) and hyperparameters. The results are shown in Table~\ref{tab:instruction_proportion} and Table~\ref{tab:rejected_action_ablation}.

Overall, if we use about half of the original dataset (in terms of both the number of instructions and the number of rejected actions), there is a drastic decrease in ACC (traj) on both of the benchmarks. Especially, in the out-of-domain benchmark, WebArena, instruction ablation results in ACC (traj) decreases from 60.0 to 15.0, which suggests it failed to generalize to unseen domains. In rejected ablation, only decreasing by one rejected action is critical, resulting 60.0 $\rightarrow$ 20.0 ACC (traj) score in WebArena. These results highlight that both the number of instructions and the number of rejected actions are critical for training an effective PRM; reducing either significantly impairs generalization, particularly in out-of-domain settings such as WebArena.

\begin{figure}[!h]
    \centering
    \includegraphics[width=0.99\linewidth]{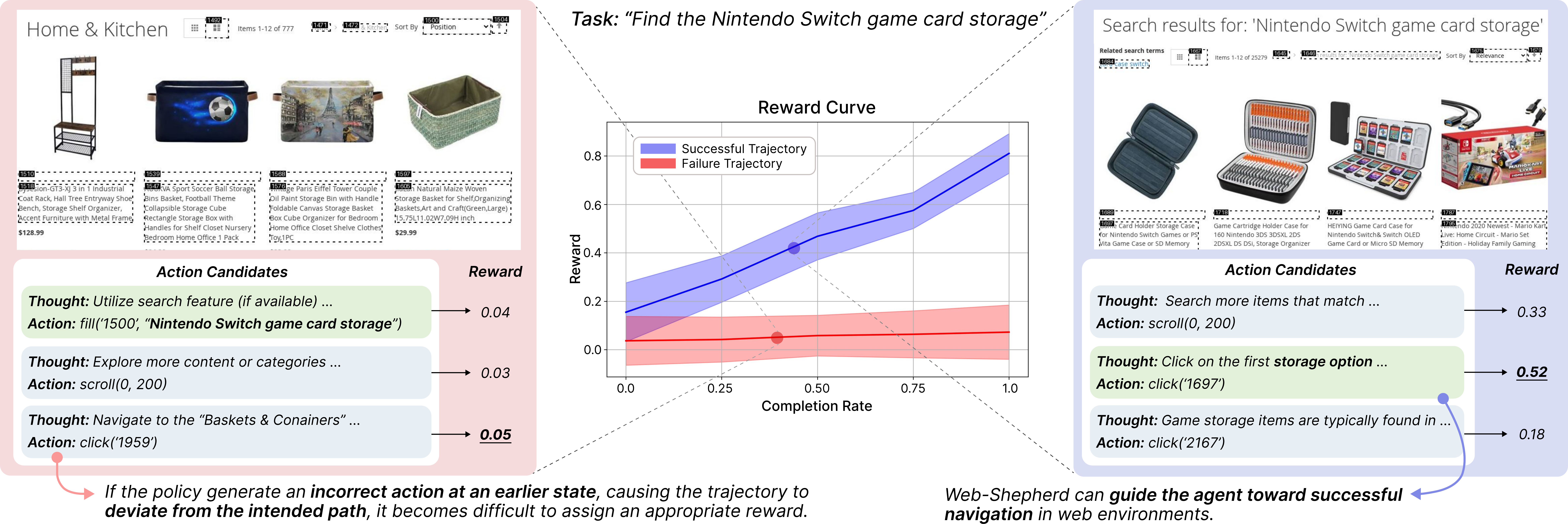}
    \caption{Trends of reward score for successful and failed cases of reward-guided trajectory search.}
    \label{fig:case_study}
\end{figure}
\subsection{Case Study}
Figure~\ref{fig:case_study} presents a qualitative analysis of \method. We sample 30 success and 30 failure cases and plot the reward score trends as a function of the normalized step index over the trajectory length. While failure cases exhibit relatively flat reward curves, successful cases show a smooth and consistent increase in reward over time. In addition, we identify the three most frequent sources of error: (1) incorrect reasoning about the effects of actions, where the model fails to anticipate future rewards appropriately—for example, assigning a low reward to a scroll action that would have revealed the desired information in the next step; (2) misinterpretation of the observed state, often due to not properly accounting for the impact of previous actions, leading the model to repeat actions unnecessarily; and (3) hallucinations in the generated checklist, such as assuming the presence of filtering functionality on a website when no such feature exists.

\section{Conclusion}
This paper studies process reward modeling for web navigation and introduces \method, the first PRM designed specifically for evaluating web agent trajectories. 
We also release two key resources to support the development of PRMs: (1) \dataset, a dataset consisting of human-annotated instructions and expert trajectories, and (2) \benchmark, a reliable benchmark designed to evaluate the capabilities of PRMs.
Our experiments demonstrate that process-level rewards improve inference-time search, achieving 34.55\% success rate on WebArena-lite compared to 23.64\% for baselines. The checklist-based approach offers a generalizable framework that could extend beyond web navigation to other sequential decision-making domains where sparse rewards and partial observability remain challenging. 
We believe \method establishes a foundation for developing more reliable web agents through interpretable reward modeling.

\section*{Acknowledgement}
This work was supported by Institute of Information \& communications Technology Planning \& Evaluation (IITP) grant funded by the Korea government (MSIT) (No. RS-2024-00457882, National AI Research Lab Project). This research was supported by the MSIT (Ministry of Science, ICT), Korea, under the Global Research Support Program in the Digital Field program (RS-2024-00436680) supervised by the IITP (Institute for Information \& Communications Technology Planning \& Evaluation). This project is supported by Microsoft Research Asia. This research was supported by a grant of Korean ARPA-H Project through the Korea Health Industry Development Institute (KHIDI), funded by the Ministry of Health \& Welfare, Republic of Korea (grant number : RS-2024-00512374). Jinyoung Yeo is the corresponding author.

\bibliographystyle{unsrtnat}
\bibliography{references}
\newpage
\appendix
\section{Limitations and Societal Impacts}
\subsection{Limitations}
\label{app:limitation}
\paragraph{Expansion to coordinate-based actions.}
Recently, coordinate-based actions—where agents interact with digital environments using direct coordinate inputs without requiring additional backend programs to convert actions—have gained attention due to their adaptability across diverse interfaces. We have also collected a dataset to extend \method to support coordinate-based action formats. However, as this direction falls outside the primary scope of this work, we leave its exploration for future research.

\paragraph{Application to reinforcement learning.}
An interesting direction for future work is to use \method as a reward signal in reinforcement learning. While we plan to explore this setting, it requires significant computational resources and is therefore left for future work. In particular, we aim to investigate whether reward signals from PRMs can improve learning efficiency—\ie how quickly rewards increase during training—as well as final performance on existing benchmarks.

\paragraph{Selection of the base model for \method.}
While our current implementation of \method uses relatively lightweight base models (3B–8B), the approach is model-agnostic and can be extended to larger scales. In principle, \method can be scaled up to stronger foundation models in the 32B–72B range, which may further improve performance in complex web environments. We leave the exploration of such scaling as future work, particularly in resource-rich settings.

\paragraph{Multimodal instructions.}
While most instructions in existing web agent benchmarks are purely textual, some tasks—such as those in VisualWebArena~\citep{koh2024visualwebarena}—incorporate both text and image modalities. Extending \method to handle multimodal instructions is a promising direction for future work, as it would enable the agent to operate in more complex and realistic web environments that require visual understanding in addition to text comprehension.

\subsection{Societal Impacts}
\label{app:societal_impacts}
\paragraph{Positive impacts.}
Web agents have the potential to perform a wide range of tasks typically carried out through a web browser, which serves as a universal interface for information access, online services, and task execution. However, current agents are often restricted to simple tasks, such as retrieving an address or clicking through static pages. We believe that \method can broaden the capabilities of web agents, enabling them to tackle more complex, goal-oriented tasks in dynamic environments. This advancement could benefit users with accessibility needs, support automated workflows in professional domains, and improve the scalability of digital assistance.

\paragraph{Negative impacts.}
Despite their potential benefits, web agents also pose several risks. Without proper safeguards, agents with the ability to autonomously interact with websites could unintentionally or maliciously perform harmful actions—such as submitting unauthorized forms, modifying user data, or accessing sensitive information. Moreover, if reward models are misaligned or insufficiently robust, agents may exploit unintended shortcuts to maximize rewards without accomplishing the intended task. To mitigate these risks, it is crucial to incorporate safety mechanisms, including strict execution constraints, permission controls, human-in-the-loop oversight, and careful auditing of model outputs in deployment scenarios.

\section{\dataset}
\label{app:wprm_collection}
\subsection{Data Annotation Toolkit}

\begin{figure}[!h]
    \centering
    \includegraphics[width=0.8\linewidth]{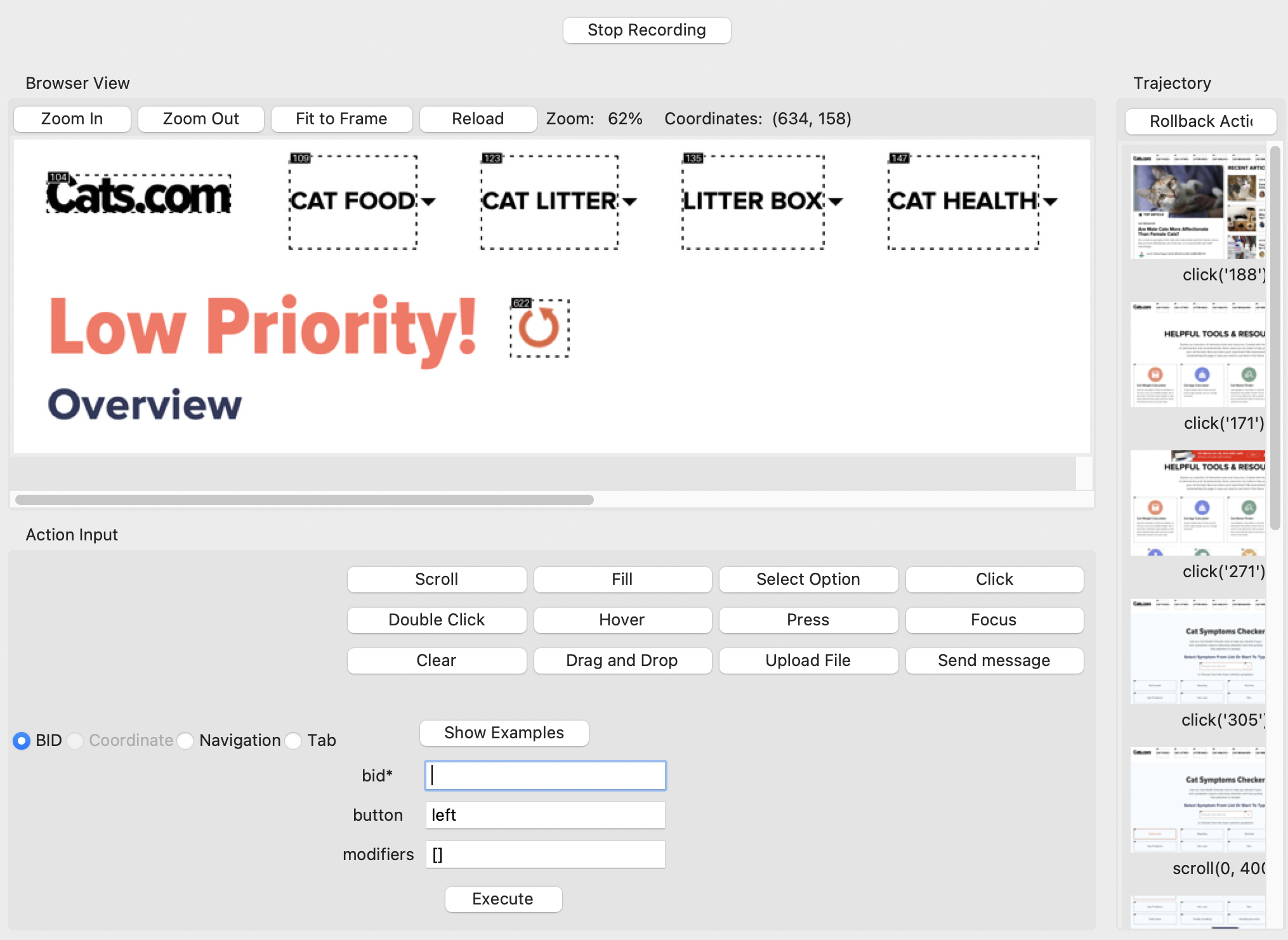}
    \caption{Our annotation toolkit used for collecting \dataset.}
    \label{fig:annotation_toolkit}
\end{figure}

To reduce the burden of human annotation, we developed a specialized toolkit for collecting web agent trajectories. It is designed to streamline the annotation process while ensuring the collection of high-quality data. Figure~\ref{fig:annotation_toolkit} shows a screenshot of the toolkit interface.
This tool helps the annotators to interact with the browser by taking the user's inputs and showing the execution output (\eg observation) within the graphical interface. 
For example, they select an action type from a predefined set (\eg \texttt{fill}) and provide the corresponding argument (\eg ``sony headphone'') via the action panel.
Selecting the action type via button clicks, rather than manually typing the entire action sequence, significantly reduces errors.
In addition, the sidebar displays snapshots of previous user interactions, allowing annotators to easily track progress, review past actions, and undo the most recent one if necessary.

\subsection{Details of Human Annotation} \label{appendix:human_annotation_details}
Our data collection process follows the steps below:

\paragraph{Step 1: Website list selection.}
We begin by selecting candidate websites from those used in the Mind2Web training dataset. Since web navigation requires browser automation, we manually filter out websites that are incompatible with our annotation process—specifically, those that block Playwright by requiring CAPTCHA verification or by rejecting HTTP requests entirely. After applying this filtering process, we retain 50 websites that are technically accessible and semantically appropriate for annotation.

\paragraph{Step 2: Annotator recruiting and education.}
We recruit groups of human annotators to construct the dataset, with overall supervision and quality control handled by designated project managers. All annotators completed a three-hour education session conducted by the project managers prior to annotation. This training covers a detailed explanation about data annotation interface, guidelines for writing quality task instructions, examples of good and bad trajectories, and principles for designing judge codes. Upon completing the training, each annotator is assigned to 3-4 websites from the filtered website list.

\paragraph{Step 3: Data annotation.}
The annotation process is structured into three distinct phases. In the first phase, we ask each annotator to create 20 task instructions for each of their assigned websites. These tasks are distributed across three difficulty levels: 5 easy, 10 medium, and 5 hard tasks. Annotators are instructed to design tasks that reflect realistic user goals, such as booking a reservation, retrieving specific information, or modifying a user profile. In the second phase, annotators execute the tasks they created and record expert trajectories interacting with our annotation toolkit (Figure~\ref{fig:annotation_toolkit}). These expert trajectories have a complete sequence of observation-action pairs needed to complete the task successfully. Lastly, annotators write a judge code that can automatically assess the trajectory towards the user's goal.
To ensure compatibility with existing benchmarks and code bases, we follow the format of judge code of WebArena~\cite{zhou2023webarena}.

\paragraph{Step 4: Verification.}
To ensure the quality of \dataset, we introduce two safeguards throughout the annotation process. 
\begin{itemize}
    \item \textbf{Automatic Verification}:
    To verify judge codes, we conduct programmatic verification that checks whether each judge code correctly evaluates the corresponding annotated trajectory as `success'. If a mismatch is detected, the annotator is instructed to revise the judge code.
    \item \textbf{Manual Verification}:
    Project managers manually review all annotated trajectories and their associated judge codes, filtering out erroneous or low-quality data. As a result, 15\% of the annotated data was discarded during this step.
\end{itemize}

\subsection{Annotating Dataset with MLLMs}
\label{app:annotation_detail_mllms}
\paragraph{Reasoning for chosen action.}
Recent web agents widely adopt a ReAct (\ie Reason + Act)~\citep{yao2023react} framework, in which the agent first produces a rationale (thought) to explain its current understanding or intent, and then selects an action based on that reasoning.
However, our human-annotated datasets lacks this intermediate reasoning step\textemdash it does not capture what the agent was thinking when choosing each action.
To enrich our dataset  with such reasoning traces, we leverage Qwen-2.5-VL-72B, prompting it with the current observation (URL, accessibility tree representation, image screenshot), the selected action and a screenshot obtained after executing the action. The model is then asked to generate a corresponding rationale that explains the  decision behind the chosen action.

\paragraph{Checklist with human trajectory.}
To effectively extract key subgoals (\ie checklist) that are essential for achieving the user's instruction, we provide \gpto with both user instruction and human trajectories, which include the intermediate thoughts.
The model is prompted to generated a reasoning process that analyzes the give task, and then to produce a checklist grounded in that reasoning.
This approach significantly enhances checklist quality.
In contrast, the low-quality checklist shown in Figure~\ref{fig:checklist_reward_relation} were generated by a model trained without the reasoning component, highlighting the importance of the task-specific reasoning in generating reliable checklists.

\paragraph{Annotating additional checklist.}
The number of checklists obtained from human trajectories is 851 instances, which is relatively small for training. To address this, we first augmented the dataset using 1K user instructions provided in the Mind2Web training set. For each instruction, we used \gpto to generate a corresponding checklist.
Subsequently, we further expanded the dataset by prompting the model to generate new instructions based on existing examples, and then constructing corresponding checklists for each, Through this augmentation process, we collected a total of 3.6K checklist instances.

\paragraph{Collecting rejected actions from various policy models.}
To construct a robust reward model, we collect diverse candidate actions from multiple policy models. These include Qwen-2.5-VL-7B and Qwen-2.5-VL-72B (both in text-only and multimodal settings), \gptomini (used specifically for generating negative actions that differ from the given chosen action), and a Qwen-2.5-3B model fine-tuned with human trajectories from \dataset. For each chosen action, we collect up to five rejected actions sampled from these policies. However, an action that differs from the chosen one is not necessarily incorrect. For example, directly filling a search box versus clicking it before typing can both be functionally valid. To eliminate such cases, we apply a rule-based filtering that retains only clearly invalid (\ie rejected) actions. 
Each action consists of a keyboard or mouse operation (\eg \texttt{click} and \texttt{fill}) and its corresponding argument, such as a unique element ID or a text string. We apply different filtering rules depending on the type of the chosen action, as detailed below:
\begin{itemize}
    \item \texttt{send\_msg\_to\_user}, \texttt{scroll}, \texttt{goto}: If the operation type differs, the candidate is considered a negative action. In particular, if the operation is \texttt{send\_msg\_to\_user}, we verify its correctness using \gpto.
    \item \texttt{drag\_and\_drop}: If the candidate action's operation is not one of \texttt{drag\_and\_drop}, \texttt{scroll}, or \texttt{hover}, it is classified as a negative action.
    \item \texttt{click}, \texttt{dclick}: If the argument (\eg element ID) does not match the chosen action’s argument and the candidate action is not semantically equivalent (\eg clicking an unrelated element), it is considered incorrect.
    \item \texttt{click}, \texttt{fill}: If both actions target the same element but differ only in order, the candidate is not considered negative. Otherwise, mismatches in target elements or unrelated inputs are marked as negative. 
    \item Others: Actions with unmatched operation types or arguments that do not lead to equivalent outcomes are treated as negative.
\end{itemize}
While this rule-based filtering substantially improves the quality of negative samples, it cannot guarantee correctness in all cases. We leave further improvement of this filtering process for future work. Finally, if more than five valid rejected actions remain after filtering, we randomly sample a subset to maintain a consistent number of action pairs per instance.

\subsection{Statistics of \dataset}
\begin{figure}[!h]
    \centering

    \begin{subfigure}[b]{\linewidth}
        \centering
        \includegraphics[width=1\linewidth]{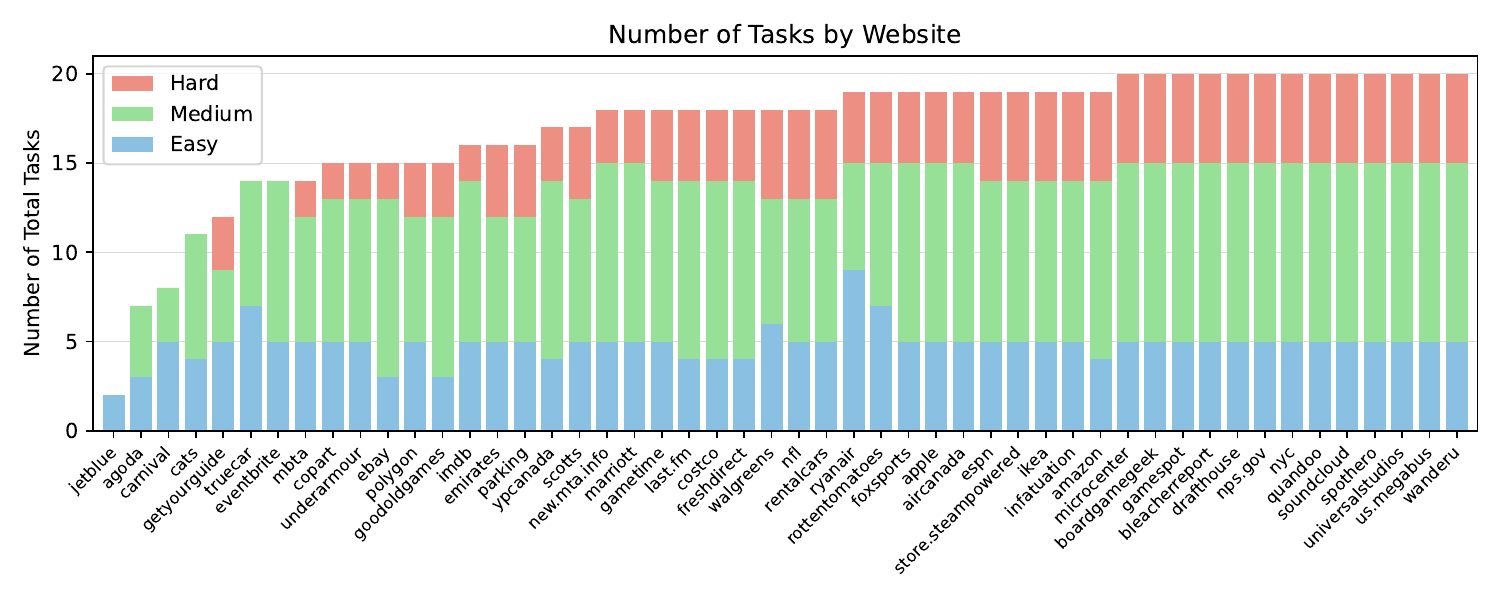}
        \caption{Number of tasks by website.}
        \label{fig:dataset_stat_num_task}
    \end{subfigure}

    \vspace{0.5em} 
    \begin{subfigure}[b]{0.4\linewidth}
        \centering
        \includegraphics[width=\linewidth]{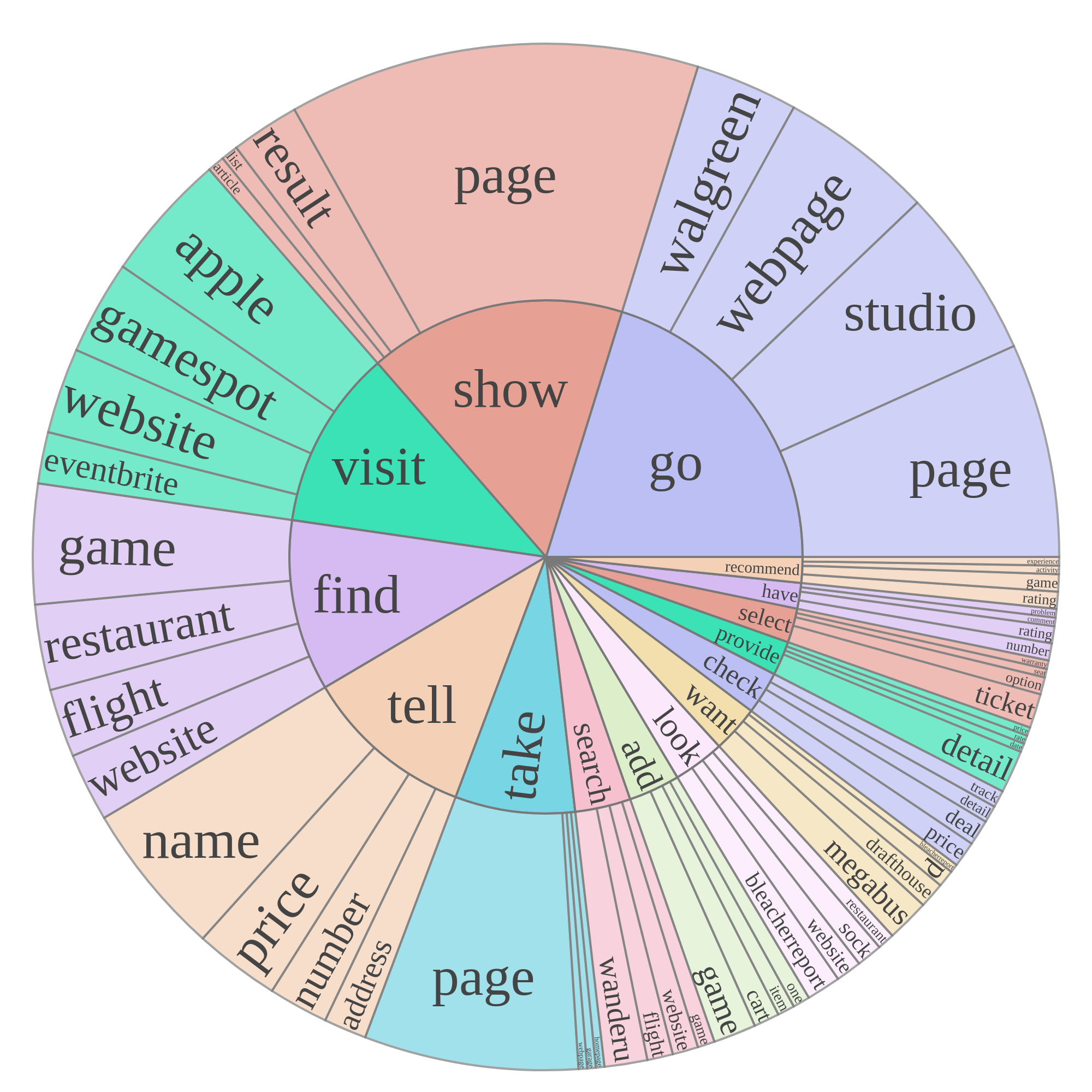}
        \caption{Instructions.}
        \label{fig:dataset_stat_instruction}
    \end{subfigure}
    \hspace{0.1\linewidth}
    \begin{subfigure}[b]{0.4\linewidth}
        \centering
        \includegraphics[width=\linewidth]{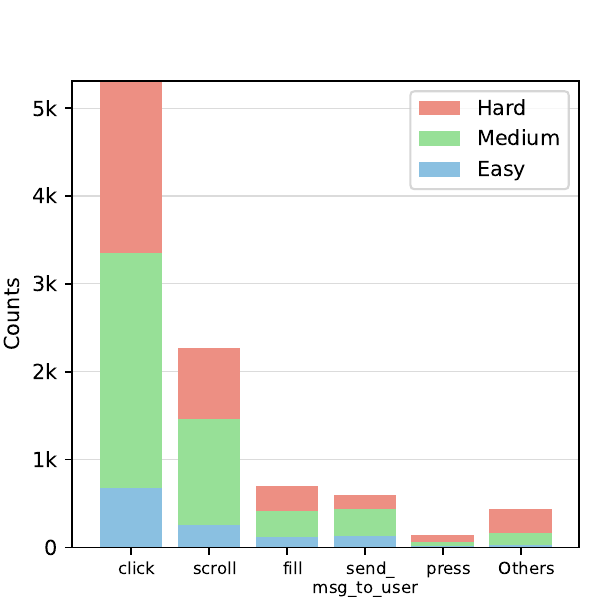}
        \caption{Actions.}
        \label{fig:dataset_stat_action}
    \end{subfigure}

    \caption{Statistics of the human-annotated dataset: (a) Number of tasks per website, grouped by difficulty (Easy, Medium, and Hard). (b) The distribution of root verbs and direct objects in instructions. (c) Action type distribution, broken down by difficulty.}
    \label{fig:dataset_stats_combined}
\end{figure}

\paragraph{Human annotated data.}
Figure~\ref{fig:dataset_stats_combined} shows the statistics of human annotated data, collected a total of 851 tasks through ad annotation process, as detailed in Appendix~\ref{appendix:human_annotation_details}. These tasks are categorized into 244 easy, 426 medium, and 181 hard tasks, covering a wide range of real-world scenarios with varying levels of complexity. Our annotated data spans a diverse set of websites, as illustrated in Figure~\ref{fig:dataset_stat_num_task}, which shows the distribution of verified tasks across different domains. A portion of annotated data\textemdash amounting of 15\%\textemdash was discarded during a manual verification step conducted by project managers to ensure data quality.

Figures~\ref{fig:dataset_stat_instruction} provides a linguistic overview of the instructions in our dataset. This sunburst chart visualizes root verbs and their most common direct objects, revealing frequent combinations such as visit webpage and find restaurant. These patterns reflect realistic user intents and highlight the diversity of task formulations in the dataset. In addition, Figure~\ref{fig:dataset_stat_action} presents the distribution of action types observed during annotation, with click, scroll, and fill appearing most frequently.

\paragraph{Rejected actions.}

\begin{figure}[!h]
    \centering
    
    \begin{subfigure}[b]{0.55\linewidth}
        \centering
        \includegraphics[width=\linewidth]{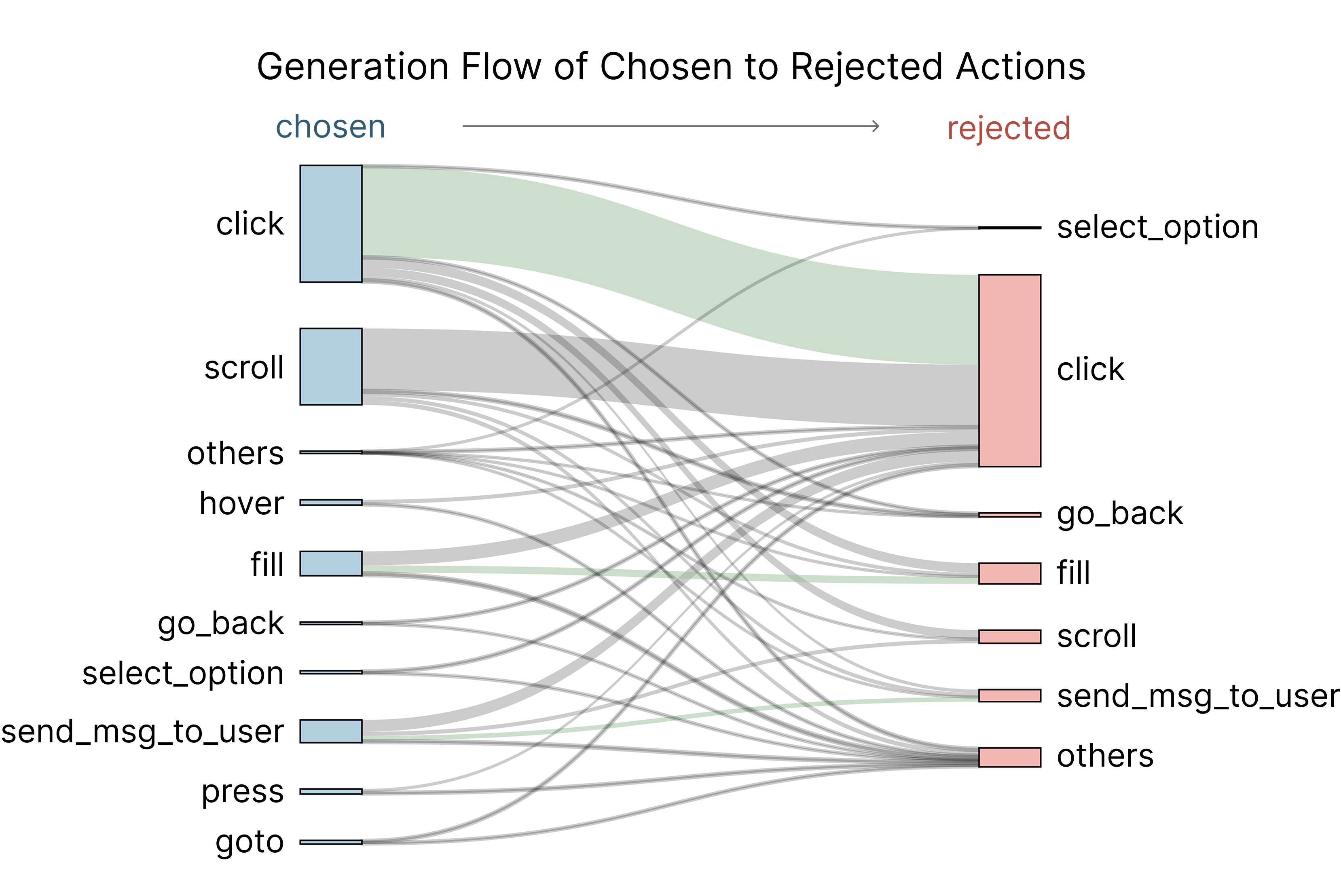}
        \caption{Generation flow of rejected actions.}
        \label{fig:rejected_stats_flow}
    \end{subfigure}
    \hfill
    \begin{subfigure}[b]{0.44\linewidth}
        \centering
        \includegraphics[width=\linewidth]{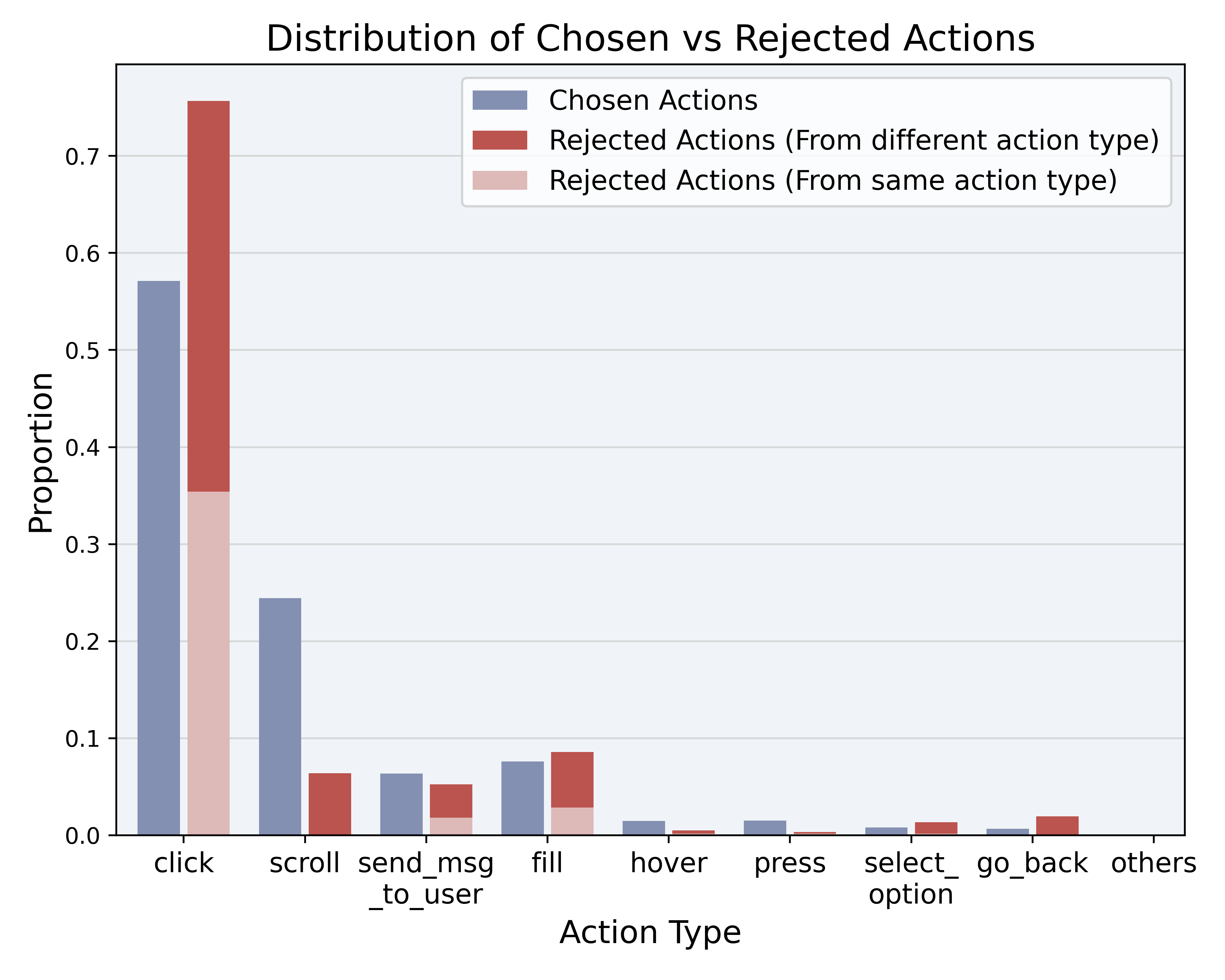}
        \caption{Proportions of chosen and rejected actions.}
        \label{fig:rejected_stats_bar}
    \end{subfigure}

    \caption{Statistics of the rejected actions: (a) Generation flow of chosen to rejected actions. Green bands indicate that the rejected actions share the same action type as their originating chosen action. (b) Proportions of chosen and rejected actions. showing how the distribution of action types shifts during the rejection generation process.}
    \label{fig:rejected_stats}
\end{figure}

After the rejected action generation step, we obtained 30,960 rejected actions from 9,473 chosen actions. Figure~\ref{fig:rejected_stats} presents an overview of the rejection statistics. The generation flow\textemdash \ie how rejected actions are derived from specific chosen actions\textemdash is shown in Figure~\ref{fig:rejected_stats_flow}. Also, Figure~\ref{fig:rejected_stats_bar} compares the distributions of chosen and rejected actions. As in the statistics, the distribution of rejected actions differs slightly from that of the chosen actions. For example, the proportion of \texttt{click} actions increased, while the proportion of \texttt{scroll} actions decreased. We leave the development of more refined methods to reduce this distributional difference to future work.

\section{\webshepherd}
\subsection{Training}

We train the model for 3 epochs with a learning rate of 1e-4, using LoRA with a rank of 16. Training is conducted using DeepSpeed ZeRO Stage 2 on an RTX A6000 (48GB) server with 8 GPUs, totaling approximately 16 GPU-hours. We leverage the LLaMA-Factory~\citep{zheng2024llamafactory} framework and apply the Liger kernel~\citep{liger2024} optimization during training.

\subsection{Inference}

We use vLLM~\citep{kwon2023efficient} to perform inference with \method. The decoding is configured with a temperature of 1.0, and nucleus sampling is applied to generate five output sequences per prompt.

To compute the probability of each label, we apply a mapping from semantic labels to token-level logits. Specifically, we aggregate the logits of the following token variants corresponding to each label:

\begin{itemize}
\item \textbf{Yes}: \texttt{["ĠYes", "Yes", "ĊYes", "Ġyes", "yes", "Ċyes", "ĠYES", "YES", "ĊYES", "ĠDone", "Done", "ĊDone", "ĠCompleted", "Completed", "ĊCompleted", "ĠCorrect", "Correct", "ĊCorrect"]}
\item \textbf{No}: \texttt{["ĠNo", "No", "ĊNo", "ĠNO", "NO", "ĊNO", "ĠNot", "Not", "ĊNot", "ĠNone", "None", "ĊNone", "ĠNope", "Nope", "ĊNope", "ĠUn", "Un", "ĊUn", "ĠWrong", "Wrong", "ĊWrong"]}
\item \textbf{In Progress}: \texttt{["ĠIn", "In", "ĊIn", "ĠPending", "Pending", "ĊPending", "ĠPart", "Part", "ĊPart", "ĠPartial", "Partial", "ĊPartial", "ĠInProgress", "InProgress", "ĊInProgress"]}
\end{itemize}

Logits corresponding to these token variants are summed for each label to compute the final label probabilities.

\label{app:webshepherd}

\section{\benchmark}
\label{app:webprmbench}

\subsection{Data Construction}
\label{app:webprmbench_detail}
\paragraph{Chosen action.}
For WebArena, we manually annotate the expert trajectories, since it is not provided in the benchmark. On the other hand, Mind2Web provide them, so we use them as the chosen actions.
One important change we make on Mind2Web is converting the HTML observation space to bid-based observation. In the HTML there exist DOM backend ids so we utilize them for the conversion.
Lastly, since it is increasingly hard to assure the quality of human-annotated rationales, we incorporate LLMs to annotate Chain-of-Thought (CoT) in a post-hoc manner.

\paragraph{Rejected actions.}
Following the setup of \citet{kim2024evaluating}, we construct a reliable benchmark by collecting multiple rejected samples from various models. In this work, we use three MLLMs\textemdash \gptomini, Qwen-2.5-VL-7B, and Qwen-2.5-VL-72B\textemdash as policy models. For each chosen action, we sample four rejected actions from these policies. To ensure that the rejected actions are truly incorrect, we apply rule-based filtering (as described in Appendix~\ref{app:annotation_detail_mllms}) and additional human filtering performed by the authors.
Finally, we collect 776 step-level data instances derived from 220 task, each associated with one chosen and four rejected actions, resulting in a total of 3,880 test instances.

\subsection{Analysis of \benchmark}
\begin{figure}[!h]
    \centering

    \begin{subfigure}[a]{0.8\linewidth}
        \centering
        \includegraphics[width=1\linewidth]{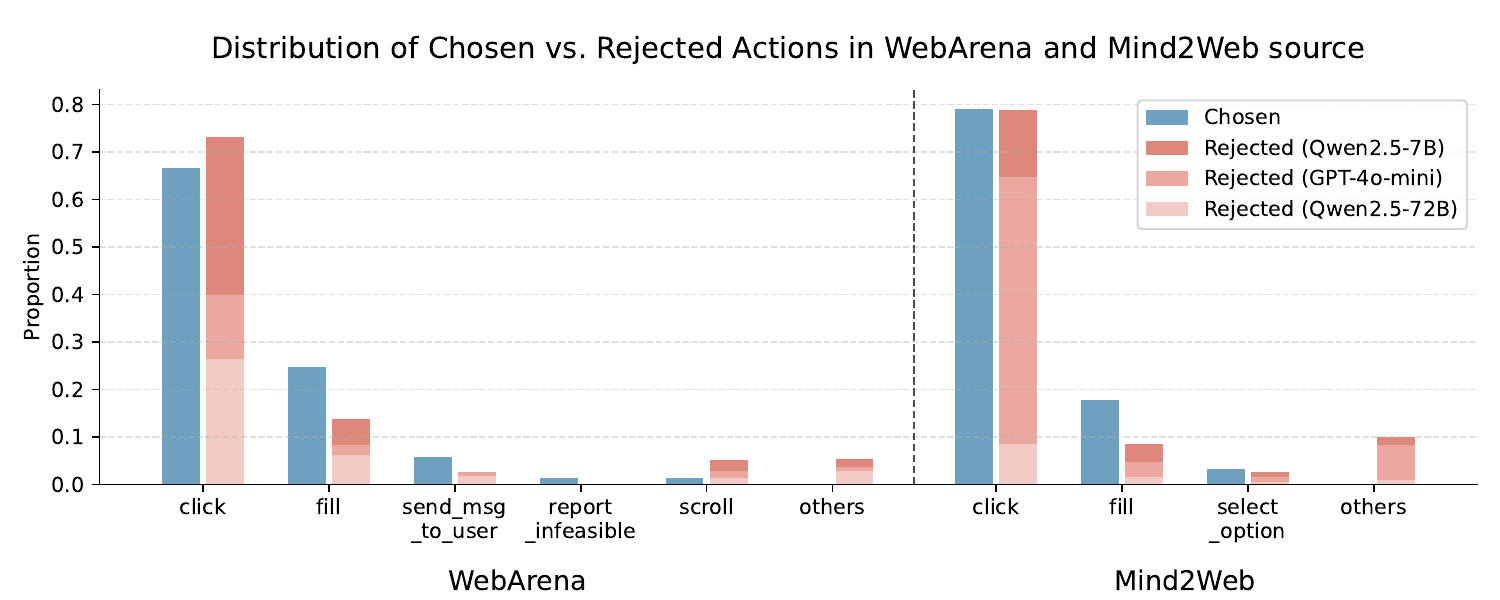}
        \caption{Distribution of chosen and rejected actions in \benchmark.}
        \label{fig:benchmark_stat_action_histogram}
    \end{subfigure}

    \vspace{0.5em} 
    \begin{subfigure}[b]{0.4\linewidth}
        \centering
        \includegraphics[width=\linewidth]{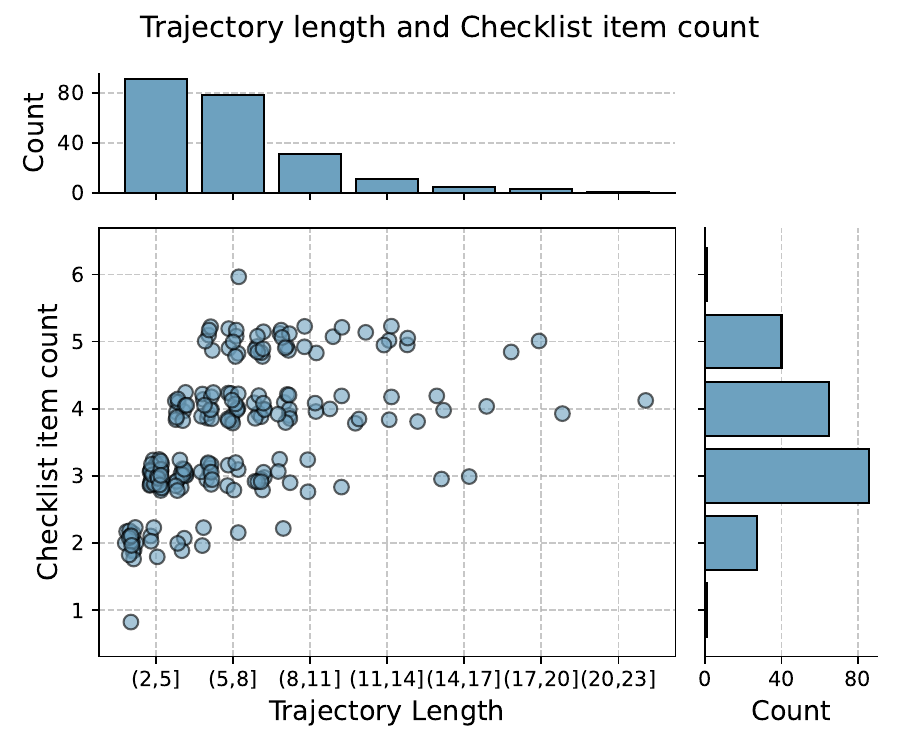}
        \caption{Trajectory length and checklist item count.}
        \label{fig:benchmark_stat_traj_checklist}
    \end{subfigure}

    \caption{Statistics of \benchmark: (a) Proportions of each chosen versus rejected action. (b) Distribution of trajectory lengths and number of checklist items.}
    \label{fig:benchmark_stat}
\end{figure}

Figure~\ref{fig:benchmark_stat_action_histogram} shows the distribution of chosen and rejected actions categorized by action type and source model across both WebArena and Mind2Web. In both datasets, \texttt{click} and \texttt{fill} actions dominate among the chosen actions, which is consistent with the typical interaction patterns required in web navigation environments. Notably, the rejected actions across all source models exhibit similar distributions, with \texttt{click} actions being the most frequently rejected. This suggests that despite differences in model architecture and scale, the failure modes of MLLMs often concentrate on similar types of actions.
Additionally, the inclusion of multiple source models—\gptomini, Qwen2.5-7B, and Qwen2.5-72B—further contributes to the diversity of rejected actions. This model-level heterogeneity ensures that the benchmark captures a broad range of suboptimal behaviors, enhancing its generality and diagnostic value.

Figure~\ref{fig:benchmark_stat_traj_checklist} visualizes the joint distribution of trajectory length and the number of checklist items associated with each task instance. The majority of trajectories fall within the 2–8 step range, while checklist items typically range from 2 to 5. The plot reveals a general trend that longer trajectories tend to be accompanied by a greater number of checklist items, indicating that tasks with longer horizons are generally more complex and goal-rich. However, we also observe several short trajectories with multiple checklist items, suggesting that brevity in execution does not necessarily imply low task complexity. This variability further highlights the importance of step-level evaluation in addition to trajectory-level metrics.

\section{Additional Results}
\label{app:additional_result}

\begin{table*}
\centering
\caption{Evaluation results on \wprmbench ~without using checklist. \textbf{T} denotes text observation, and \textbf{I} denotes image observation. Acc. (s) refers to step accuracy, while Acc. (t) refers to trajectory accuracy.}

\resizebox{0.99\textwidth}{!}{
\begin{tabular}{lcccccccccccccc}
\toprule
\multirow{3}{*}{\textbf{Model}} & \multirow{3}{*}{\textbf{Inputs}} & \multicolumn{9}{c}{\textbf{Mind2Web}} & \multicolumn{3}{c}{\textbf{WebArena}}
\\
\cmidrule(lr){3-11}\cmidrule(lr){12-14}  
&  &  \multicolumn{3}{c}{\textit{Cross-Task}} & \multicolumn{3}{c}{\textit{Cross-Website}} & \multicolumn{3}{c}{\textit{Cross-Domain}} & \multicolumn{3}{c}{\textit{Test}} \\ 
& & MRR  & $\text{Acc. (s)}$ & $\text{Acc. (t)}$ & MRR  & $\text{Acc. (s)}$ & $\text{Acc. (t)}$ & MRR   & $\text{Acc. (s)}$& $\text{Acc. (t)}$ & MRR  & $\text{Acc. (s)}$ & $\text{Acc. (t)}$\\
\midrule
\multicolumn{14}{c}{\cellcolor{gray!11}\textit{Reward Assignment with \textbf{Likert Scale}}}  \\
\midrule
\multirow{2}{*}{\gptomini}  &  \textbf{T} & 47.5 & 15.5 & 0.0 & 47.6 & 13.5 & 0.0 & 45.4 & 11.8 & 0.8 & 34.4 & 5.8 & 5.0 \\
& \textbf{T} + \textbf{I} & 44.7 & 12.7 & 2.5 & 42.8 & 8.8 & 0.0 & 43.1 & 10.1 & 0.0 & 34.6 & 8.7 & 5.0 \\
\midrule
\multirow{2}{*}{\gpto} & \textbf{T} & 56.9 & 28.8 & 5.0 & 55.8 & 26.4 & 2.6 & 59.8 & 33.6 & 3.3 & 59.2 & 37.7 & 15.0 \\
& \textbf{T} + \textbf{I} & 52.5 & 21.8 & 5.0 & 52.2 & 21.0 & 0.0 & 52.8 & 23.3 & 1.7 & 50.0 & 24.6 & 5.0 \\
\midrule
\multirow{2}{*}{\qwenvllarge} & \textbf{T} & 55.7 & 26.1 & 5.0 & 51.8 & 20.3 & 0.0 & 54.2 & 24.7 & 1.7 & 54.6 & 31.9 & 5.0 \\
& \textbf{T} + \textbf{I} & 53.5 & 23.2 & 2.5 & 47.6 & 15.5 & 0.0 & 49.8 & 19.4 & 0.8 & 43.1 & 15.9 & 0.0 \\
\midrule
\multicolumn{14}{c}{\cellcolor{gray!11}\textit{Reward Assignment with \textbf{3 Class}}}  \\
\midrule
GPT-4o-mini & \textbf{T} & 44.7 & 12.7 & 2.5 & 42.8 & 8.8 & 0.0 & 43.1 & 10.1 & 0.0 & 34.6 & 8.7 & 5.0 \\
\midrule
GPT-4o & \textbf{T} & 49.3 & 17.6 & 2.5 & 44.5 & 12.2 & 2.6 & 47.2 & 16.6 & 0.0 & 44.9 & 20.3 & 0.0 \\
\midrule
Qwen-2.5-VL-72B & \textbf{T} & 50.6 & 25.4 & 7.5 & 53.3 & 29.1 & 5.1 & 54.4 & 30.5 & 2.5 & 48.0 & 24.6 & 0.0 \\
\bottomrule
\end{tabular}
}
\label{tab:app_prmbench_with_checklist}
\end{table*}
\begin{table*}
\centering
\caption{Evaluation results on \wprmbench ~with using checklist. Results are averaged over four test set types, and reflect performance under different setting, including whether the ``In Progress'' label is used during prediction.}

\resizebox{0.55\textwidth}{!}{
\begin{tabular}{lcccccc}
\toprule
\multirow{2}{*}{\textbf{Model}} & \multirow{2}{*}{\textbf{Use `In Progress'}} & \multicolumn{3}{c}{\wprmbench }\\
\cmidrule(lr){3-5}  
& & MRR  & $\text{Acc. (step)}$ & $\text{Acc. (traj)}$ \\
\midrule
\multirow{2}{*}{GPT-4o-mini}  & \redx & 59.3 & 33.5 & 5.2 \\
 & \greencheck & 63.3 & 40.7 & 11.3 \\
\bottomrule
\end{tabular}
}
\label{tab:app_in_progress}
\end{table*}

\begin{table*}
\centering
\caption{Evaluation results on \wprmbench with using checklist. \textbf{T} denotes text observation, and \textbf{I} denotes image observation. Acc. (s) refers to step accuracy, while Acc. (t) refers to trajectory accuracy.}

\resizebox{0.99\textwidth}{!}{
\begin{tabular}{lcccccccccccccc}
\toprule
\multirow{3}{*}{\textbf{Model}} & \multirow{3}{*}{\textbf{Inputs}} & \multicolumn{9}{c}{\textbf{Mind2Web}} & \multicolumn{3}{c}{\textbf{WebArena}}
\\
\cmidrule(lr){3-11}\cmidrule(lr){12-14}  
&  &  \multicolumn{3}{c}{\textit{Cross-Task}} & \multicolumn{3}{c}{\textit{Cross-Website}} & \multicolumn{3}{c}{\textit{Cross-Domain}} & \multicolumn{3}{c}{\textit{Test}} \\ 
& & MRR  & $\text{Acc. (s)}$ & $\text{Acc. (t)}$ & MRR  & $\text{Acc. (s)}$ & $\text{Acc. (t)}$ & MRR   & $\text{Acc. (s)}$& $\text{Acc. (t)}$ & MRR  & $\text{Acc. (s)}$ & $\text{Acc. (t)}$\\
\midrule
\multicolumn{14}{c}{\cellcolor{gray!11}\textit{Reward Assignment with \textbf{Reference Checklist}}}  \\
\midrule
\multirow{2}{*}{\gptomini}  &  \textbf{T} & 63.9 & 40.1 & 5.0 & 66.1 & 42.6 & 5.0  & 63.3 & 40.8 & 12.4 & 60.0 & 39.1 & 15.0\\
& \textbf{T} + \textbf{I} & 58.8 & 33.8 & 2.5 & 64.4 & 41.2 & 5.1 & 63.0 & 39.3 & 4.1 & 53.8 & 29.0 & 5.0 \\
\midrule
\multirow{2}{*}{\gpto} & \textbf{T} & 67.4 & 46.5 & 7.5 & 70.3 & 52.0 & 5.1 & 70.2 & 51.3 & 11.6 & 69.7 & 53.6 & 15.0 \\
& \textbf{T} + \textbf{I} & 62.4 & 39.4 & 5.0 & 68.1 & 50.0 & 15.4 & 65.1 & 43.2 & 6.6 & 60.0 & 37.7 & 10.0 \\
\midrule
\multirow{2}{*}{\qwenvllarge} & \textbf{T} & 59.4 & 35.2 & 0.0 & 62.4 & 40.5 & 0.0 & 57.9 & 32.9 & 1.7 & 52.3 & 30.4 & 5.0 \\
& \textbf{T} + \textbf{I} & 52.9 & 28.2 & 2.5 & 53.5 & 27.7 & 2.6 & 52.0 & 25.9 & 2.5 & 47.3 & 24.6 & 0.0 \\
\midrule
Claude-3.7-sonnet & \textbf{T} & 60.7 & 41.6 & 7.5 & 58.7 & 34.5 & 10.3 & 60.3 & 40.3 & 5.8 & 55.2 & 37.7 & 5.0 \\
\midrule
Gemini-2.5-flash & \textbf{T} & 53.4 & 27.5 & 5.0 & 59.7 & 35.8 & 7.7 & 57.2 & 32.1 & 4.1 & 57.2 & 36.2 & 0.0 \\
\midrule
\multirow{2}{*}{\method (3B)} & \textbf{T} & 87.6 & 80.3 & 55.0 & 88.0 & 79.7 & 43.6 & 87.2 & 79.1 & 47.1 & 91.1 & 85.5 & 60.0 \\
& \textbf{T} + \textbf{I} & 85.0 & 76.8 & 42.5 & 87.3 & 79.1 & 41.0 & 84.4 & 74.1 & 37.2 & 92.5 & 87.0 & 65.0 \\
\midrule
\method (8B)& \textbf{T} & 88.8 & 82.4 & 57.5 & 87.9 & 80.4 & 51.3 & 91.3 & 85.9 & 61.2 & 97.8 & 95.7 & 85.0 \\
\midrule
\multicolumn{14}{c}{\cellcolor{gray!11}\textit{Reward Assignment with \textbf{Checklist Generation}}}  \\
\midrule
\multirow{2}{*}{\gptomini}  &  \textbf{T} & 55.3 & 30.3 & 2.5 & 57.7 & 29.7 & 5.1 & 56.6 & 31.7 & 4.1 & 51.3 & 30.4 & 5.0 \\
& \textbf{T} + \textbf{I} & 59.9 & 36.6 & 2.5 & 55.4 & 27.0 & 0.0 & 57.0 & 32.6 & 5.0 & 57.8 & 37.7 & 15.0 \\
\midrule
\gpto & \textbf{T} & 59.6 & 39.4 & 7.5 & 54.8 & 32.4 & 2.6 & 58.4 & 36.9 & 4.1 & 55.4 & 34.8 & 5.0 \\
\midrule
\qwenvllarge & \textbf{T} & 50.4 & 23.2 & 2.5 & 54.8 & 28.4 & 0.0 & 54.8 & 29.0 & 2.5 & 52.4 & 31.9 & 0.0 \\
\midrule
\multirow{2}{*}{\method (3B)} & \textbf{T} & 85.3 & 75.4 & 50.0 & 83.8 & 74.3 & 33.3 & 84.8 & 75.3 & 39.7 & 94.6 & 89.9 & 70.0 \\
& \textbf{T} + \textbf{I} & 81.1 & 69.7 & 25.0 & 78.6 & 64.9 & 23.1 & 77.9 & 64.3 & 22.3 & 85.9 & 75.4 & 40.0 \\
\midrule
\method (8B)& \textbf{T} & 87.3 & 80.3 & 50.0 & 84.3 & 76.4 & 38.5 & 86.0 & 76.7 & 43.8 & 96.5 & 94.2 & 80.0 \\
\bottomrule
\end{tabular}
}
\label{tab:app_prmbench_wo_checklist}
\end{table*}
\subsection{Evaluating MLLMs as Process Reward Models for Web Navigation}
We conduct experiments to investigate the most suitable format for reward prediction when using MLLMs as preference reward models (PRMs). Specifically, we evaluate how helpful a generated action is in progressing toward the goal from the current state.
We consider two formats: a Likert scale rating (1-5) and a 3-class classification with labels \textit{helpful}, \textit{neutral}, \textit{not helpful}. To reduce variance, each instance is sampled five times and the scores are averaged.
Table~\ref{tab:app_prmbench_wo_checklist} shows that the Likert scale consistently outperforms the 3-class classification, indicating that fine-grained evaluation provides a more informative learning signals.

Furthermore, we examine how reward prediction changes when a reference checklist is provided. We compare two evaluation schemes: one that uses binary labels (`Yes' or `No') for each checklist item, and another that introduces an additional label (`In Progress') to indicate when an action partially completes a checklist item.
As shown in Table~\ref{tab:app_in_progress}, incorporating the \textit{In Progress} label leads to more reliable reward assignments when using checklists. Based on this finding, we adopt the \textit{In Progress} setting for checklist-based reward prediction in both the \benchmark evaluation and the training of \method.

\subsection{Detailed Results of Refinement}
\begin{table*}
\centering
\caption{Detailed results of refinement with feedback from \webshepherd in Table~\ref{tab:reflexion}}
\resizebox{0.90\textwidth}{!}{
\begin{tabular}{ll|ccccc|c|c}
\toprule
\textbf{Policy} & \textbf{Model} & \textbf{Shopping} & \textbf{CMS} & \textbf{Reddit} & \textbf{GitLab} & \textbf{Map} & \textbf{Total} & \textbf{$\Delta$} \\ 
\midrule
\multirow{3}{*}{GPT-4o-mini}  & w/o refine & 21.74 & 22.86 & 19.05 & \textbf{34.38}& 19.35 &  23.64 & -- \\
  & 
    \method (3B)& \textbf{23.91}& 31.43& 19.05& \textbf{34.38}& \textbf{22.58}& 26.67& $+3.03$\\
  &  \method (8B)& \textbf{23.91}& \textbf{34.29}& \textbf{33.33}& \textbf{34.38}& 16.13& \textbf{27.88}& \textbf{$+4.24$}\\
      \bottomrule
\end{tabular}
}
\label{tab:app_reflexion}
\end{table*}

We conduct experiments for refinement with feedback from reward models (Table~\ref{tab:reflexion}). We show the detailed experimental results in Table~\ref{tab:app_reflexion}.

\subsection{Scoring Strategy: Probability vs. Token}
\label{app:scoring_strategy}
\begin{table*}
\centering
\caption{The impact of scoring strategies on reward assignment. Results are evaluated on \wprmbench and represent the average score across four types of test sets.}

\resizebox{0.65\textwidth}{!}{
\begin{tabular}{lcccccc}
\toprule
\multirow{2}{*}{\textbf{Model}} & \multirow{2}{*}{\textbf{Strategy}} & \multicolumn{3}{c}{\wprmbench }\\
\cmidrule(lr){3-5}  
& & MRR  & $\text{Acc. (step)}$ & $\text{Acc. (traj)}$ \\
\midrule
\multirow{4}{*}{\method (3B)}  & 1 res & 72.7 & 60.7 & 12.2 \\
 & 1 prob & 86.3 & 77.0 & 43.1 \\
 & 5 avg & 83.7 & 75.2 & 29.5 \\
 & 5 prob & \textbf{88.5} & \textbf{81.2} & \textbf{51.4}  \\ 
\midrule
\multirow{4}{*}{\method (8B)}  & 1 res & 77.7 & 67.4 & 14.8  \\
 & 1 prob & 86.9 & 79.2 & 48.2 \\
 & 5 avg & 88.8 & 82.9 & 51.6 \\
 & 5 prob & \textbf{91.3} &\textbf{ 86.1} & \textbf{63.7} \\ 
\bottomrule
\end{tabular}
}
\label{tab:app_scoring_strategy}
\end{table*}

When using generative models for reward prediction, on can either directly interpret the model's natural language output (\eg `Yes') as a reward signal or compute the probability of specific response~\cite{zhang2024generative} to derive a reward. To investigate which approach is more effective, we compare the following strategies:
\begin{itemize}
    \item \textbf{1 res}: Sample a single response at temperature 0 and use the output directly for reward assignment.
    \item \textbf{1 prob}: Compute the probability of a specific word (\eg `Yes') at temperature 0.
    \item \textbf{5 avg}: Sample five responses at temperature 1, convert each to a reward directly and compute average.
    \item \textbf{5 prob}: Sample five responses at temperature 1 and compute the average probability of the targe word.
\end{itemize}
This setup allows us to analyze the trade-offs between deterministic and stochastic decoding, as well as between output-based and probability-based reward estimation.
As illustrated in Table~\ref{tab:app_scoring_strategy}, we observe that sampling multiple responses (\eg 5 samples) leads to more effective reward estimation overall. When using only a single sample, computing the probability or the target tokens yeilds significantly better results than relying on the raw token output\textemdash especially at the treajectory level, where the performance gap is more pronounced.

\subsection{Relationship between Reward and Task Success}
\begin{figure}[!h]
    \centering
    \includegraphics[width=1\linewidth]{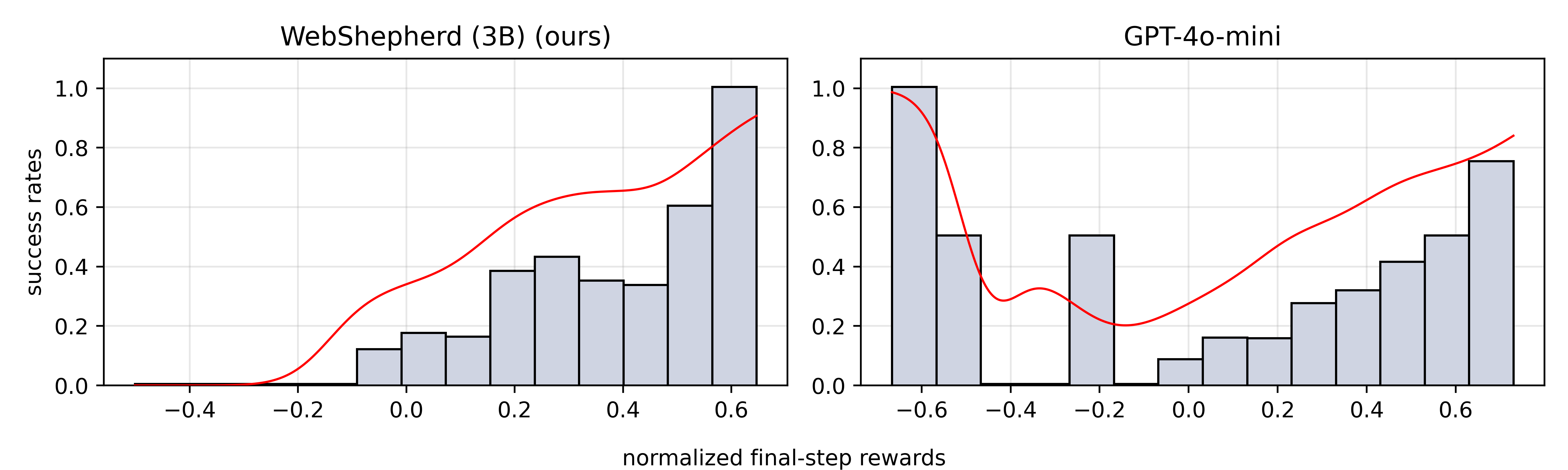}

    \caption{Task success rates binned by normalized final-step reward for \webshepherd (3B) (left) and GPT-4o-mini (right).}
    \label{fig:reward_success_alignment}
\end{figure}

A potential issue in using signal from reward model in RL is \textit{reward over-optimization}, where policy is overfitted to the imperfect reward signal~\cite{kim2024evaluating}. In such cases, the model may receive high reward signals despite failing the actual task, resulting in degraded performance.
To mitigate this, the reward model must be well-aligned with actual task success and progression.
Therefore, we examine the alignment between \webshepherd and task success. Figure~\ref{fig:reward_success_alignment} presents the correlation between final-step rewards and task success, based on the reward-guided trajectory search results described in Section~\ref{ssec:BoN}. To compare rewards across trajectories, we normalize the final-step reward by subtracting the average reward of preceding steps within the same trajectory. For \webshepherd, we observe that higher normalized final-step rewards are associated with higher success rates, while \gptomini shows no clear correlation between normalized rewards and task success. This suggests that \webshepherd is better aligned with actual task success and thus less susceptible to reward over-optimization compared to \gptomini.

\begin{table*}
\centering
\caption{Impact of the ratio between chosen and rejected samples on \webshepherd's performance.}

\resizebox{0.9\textwidth}{!}{
\begin{tabular}{lc|c|ccccc}
\toprule
\multirow{2}{*}{\textbf{Model}} & \multirow{2}{*}{\shortstack{\textbf{Sample ratio}\\\textbf{(chosen : rejected)}}} & \multicolumn{6}{c}{WebArena-Lite~\cite{liu2025visualagentbench}} \\
\cmidrule(lr){3-8}  
& & \textbf{Total} & Shopping & CMS & Reddit & GitLab & Map \\
\midrule
\multirow{2}{*}{\method (8B)}  
& $\text{1 : 1}$  & 24.85 & 23.91 & 25.71 & \textbf{19.05} & \textbf{34.38} & 19.35 \\
& $\text{1 : 4 (Ours)}$ & \textbf{32.12} & \textbf{32.61} &\textbf{ 37.14} & \textbf{19.05} & \textbf{34.38} & \textbf{32.26} \\
\bottomrule
\end{tabular}
}
\label{tab:app_ratio_exp}
\end{table*}

\subsection{The Impact of the Ratio between Chosen and Rejected Actions in Training Dataset}
To better understand the effect of learning to criticize rejected actions, we construct training datasets with two different ratios of positive to negative examples: 1:1 and 1:4. Using GPT-4o-mini as the policy model, we conduct trajectory search experiments on WebArena-Lite.
As shown in Table~\ref{tab:app_ratio_exp}, models trained with the 1:4 ratio provide more effective guidance at inference-time. This finding suggests that learning to predict rewards across diverse set of rejected actions is more beneficial, even wen the ratio of positives is highly imbalanced.

\subsection{Cost Efficiency}
\begin{wraptable}[9]{r}{0.4\textwidth}
\centering
\vspace{-2em}
\caption{Cost per 1,000 instances (USD) across different models.}
\resizebox{0.32\textwidth}{!}{
\begin{tabular}{lrr}
\toprule
\textbf{Model} & \textbf{Cost (USD)} \\
\midrule
GPT-4o & 435.74 \\
GPT-4o-mini & 43.57 \\
Qwen-2.5-VL-72B & 53.69 \\
Claude Sonnet 3.7 & 273.16 \\
Gemini 1.5 (Pro) & 13.37 \\
\midrule
\method\-3B & 4.67 \\
\bottomrule
\end{tabular}
}
\vspace{5em}
\label{tab:app_cost}
\end{wraptable}
We provide the full cost breakdown for \method and the baseline models in Table~\ref{tab:app_cost}. The cost of \method is estimated as follows: first, we compute the number of input and output tokens per instance by running the model on the evaluation set. Then, we measure the throughput—defined as the number of tokens (input + output) processed per minute—on a server equipped with a single A100 80GB GPU. Finally, using the hourly cost of the hardware (\$1.19/hour), we derive the cost per 1,000 instances. This allows for a fair comparison with API-based models, whose costs are based on pricing information from OpenRouter\footnote{\url{https://openrouter.ai/qwen/qwen2.5-vl-72b-instruct}}, OpenAI\footnote{\url{https://openai.com/pricing}}, Anthropic\footnote{\url{https://www.anthropic.com/pricing}}, and Google\footnote{\url{https://ai.google.dev/gemini-api/docs/pricing}}.

\subsection{Evaluating the Quality of the Generated Checklist}
\label{app:checklist_quality}
We use LLM-as-a-Judge method~\citep{liu2023g_eval} to evaluate the quality of the generated checklists. However, since LLMs are trained on large-scale web data but do not possess complete or up-to-date knowledge of all websites, their evaluations can be unreliable in this context. To address this limitation, we provide a reference checklist during evaluation, allowing the LLM to assess the generated output relative to a known, task-specific ground truth. We evaluate the quality of checklist along three key dimensions: (1) \textbf{Validity}\textemdash whether any incorrect or irrelevant checklist items are generated; (2) \textbf{Subgoal Granularity}\textemdash whether the steps are overly fine-grained or unnecessarily detailed; and (3) \textbf{Goal Coverage}\textemdash whether the checklist includes all key steps necessary to complete the final goal.
Specifically, the LLM is prompted to assign a quality score on a Likert scale (\ie from 1 to 5), along with a rationale explaining the evaluation. To reduce evaluation variance, each instance is rated three times, and we report the average score.
\begin{table*}
\centering
\caption{Results of G-Eval on checklist quality evaluation.}

\resizebox{0.85\textwidth}{!}{
\begin{tabular}{lc|ccccc}
\toprule
\textbf{Checklist Source} & \textbf{Overall} & \textbf{Validity} & \textbf{Subgoal Granularity}& \textbf{Goal Coverage}  \\
\midrule
GPT-4o-mini & 4.14 & 4.39  & 4.12 & 3.92  \\
GPT-4o & 4.16 & 4.42 & 4.15 & 3.90 \\
Qwen-2.5-VL-72B& 4.18 & 4.36 & \textbf{4.21} & \textbf{3.97} \\
Early version of ours (low quality) & 2.97 & 3.08 & 3.10 & 2.74 \\
\method (3B)& 3.86 & 4.00 & 3.97 & 3.60 \\
\method (8B) & 3.91 & 4.07 & 3.97 & 3.68  \\
Checklist only model (ours, 8B) & \textbf{4.21} & \textbf{4.50} & 4.17 & \textbf{3.97} \\
\bottomrule
\end{tabular}
}
\label{tab:app_g_eval}
\end{table*}

Table~\ref{tab:app_g_eval} presents the checklist quality across different checklist sources, evaluated along three dimensions and an overall score. We observe that, with the exception of our initial model version (trained only on checklist generation without reasoning) and \method (3B), most models produce checklists of comparable quality. Notably, the model trained solely for checklist generation (\ie without multi-task learning like \method), suggesting the benefit of task-specific supervision. Based on this observation, we also release the checklist\textemdash only model to support broader use cases.

\section{Details of Experiments}
\subsection{\benchmark}
\paragraph{Evaluation.}
We evaluate model performance under the following default setting: five sampled outputs are generated using a temperature of 1.0. In the baseline setting without a checklist, outputs are assessed using a Likert-scale. For the checklist-based reward prediction setup, we evaluate the completion status of each checklist item using three labels: \textit{Yes}, \textit{In Progress}, and \textit{No}.
The prompts used to evaluate PRMs on \benchmark are presented below:
\begin{itemize}
    \item Reward prediction w/o checklist: Figure~\ref{fig:prompt_likert}
    \item Checklist generation: Figure~\ref{fig:prompt_checklist_generation} (baseline), Figure~\ref{fig:prompt_checklist_generation_ours} (ours)
    \item Reward prediction based on checklist: Figure~\ref{fig:prompt_checklist_reward} (baseline), Figure~\ref{fig:prompt_reward_ours} (ours)
\end{itemize}

\subsection{Reward-guided Trajectory Search}
\paragraph{Environment.}
We use BrowserGym~\citep{chezelles2024browsergym}, a unified framework for evaluating web agents in online environments. BrowserGym standardizes the action space across different implementations, improving reproducibility. It also processes both textual and visual observations through an overlay of set-of-marks, enabling richer interaction signals. Additionally, it supports automatic Docker-based website resets and identifies task dependencies to prevent unintended side effects between tasks.

\paragraph{Policy and action selection.}
We use \gptomini and \gpto as the policy model. To obtain action candidates, we sample 20 output sequences using nucleus sampling with a temperature of 1.0. The top-$n$ most frequent actions across these samples are selected as candidates. 

We then score each candidate action using the reward model and select the one with the highest predicted reward. In cases where multiple actions receive the same score, we execute the action that was sampled more frequently.

\paragraph{Refinment experiments.}
We use reward model's thought and checklist evaluation responses\textemdash excluding the actual reward score\textemdash as feedback for refinement. The refinement is repeated up to two times, as long as it leads to a higher reward score than the previous step. At the end of the refinement step, we obtain up to three action candidates and select the one with the highest reward score as the final action.

\section{Case Study}
\begin{figure}[!h]
    \centering
    \includegraphics[width=1.0\linewidth]{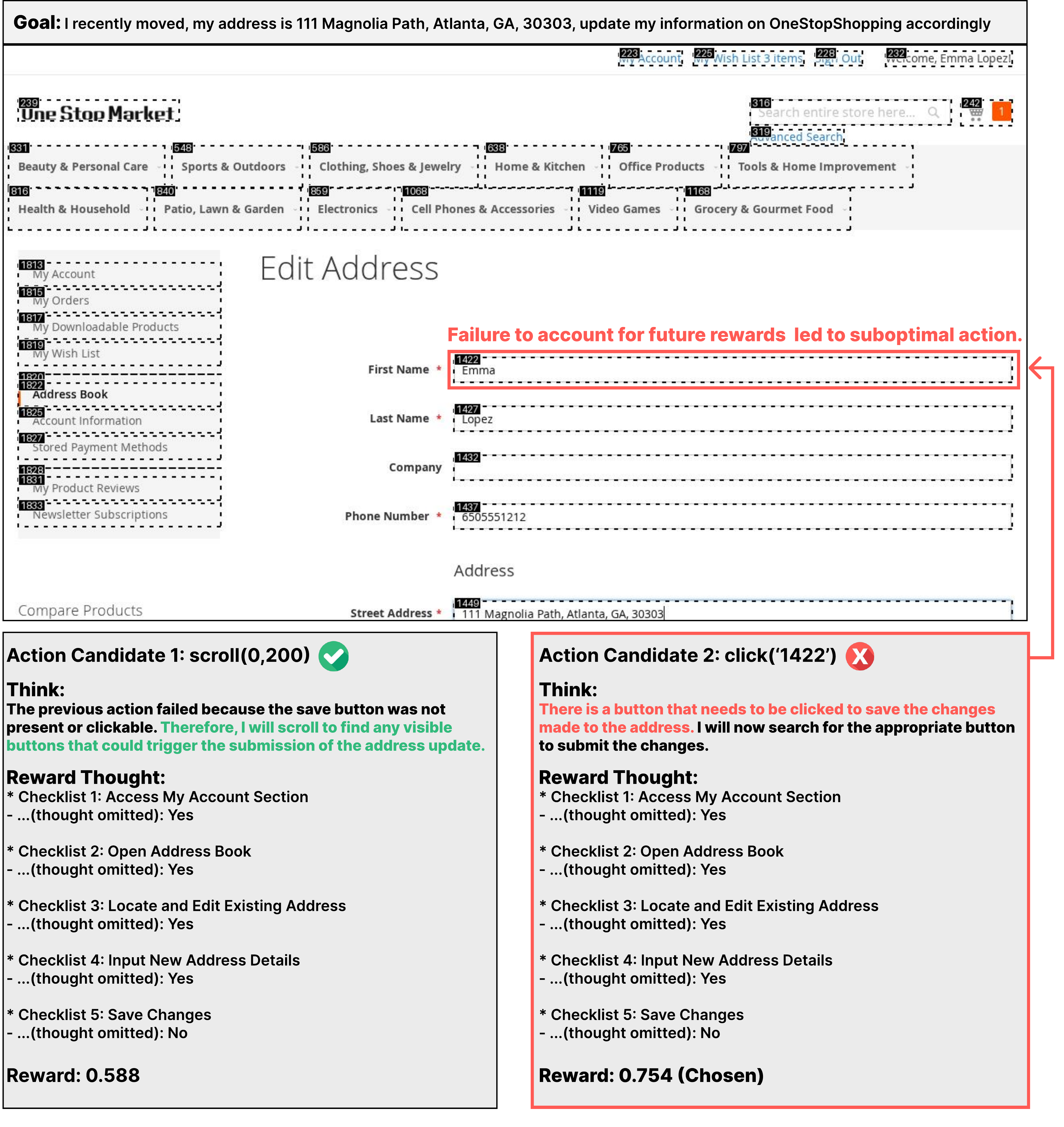}
    \caption{Erroneous Example. Failed to properly anticipate future rewards, the agents clicked a sub-optimal bid instead of scrolling to find the save button.}
    \label{fig:case_study_1}
\end{figure}
\begin{figure}[!h]
    \centering
    \includegraphics[width=1.0\linewidth]{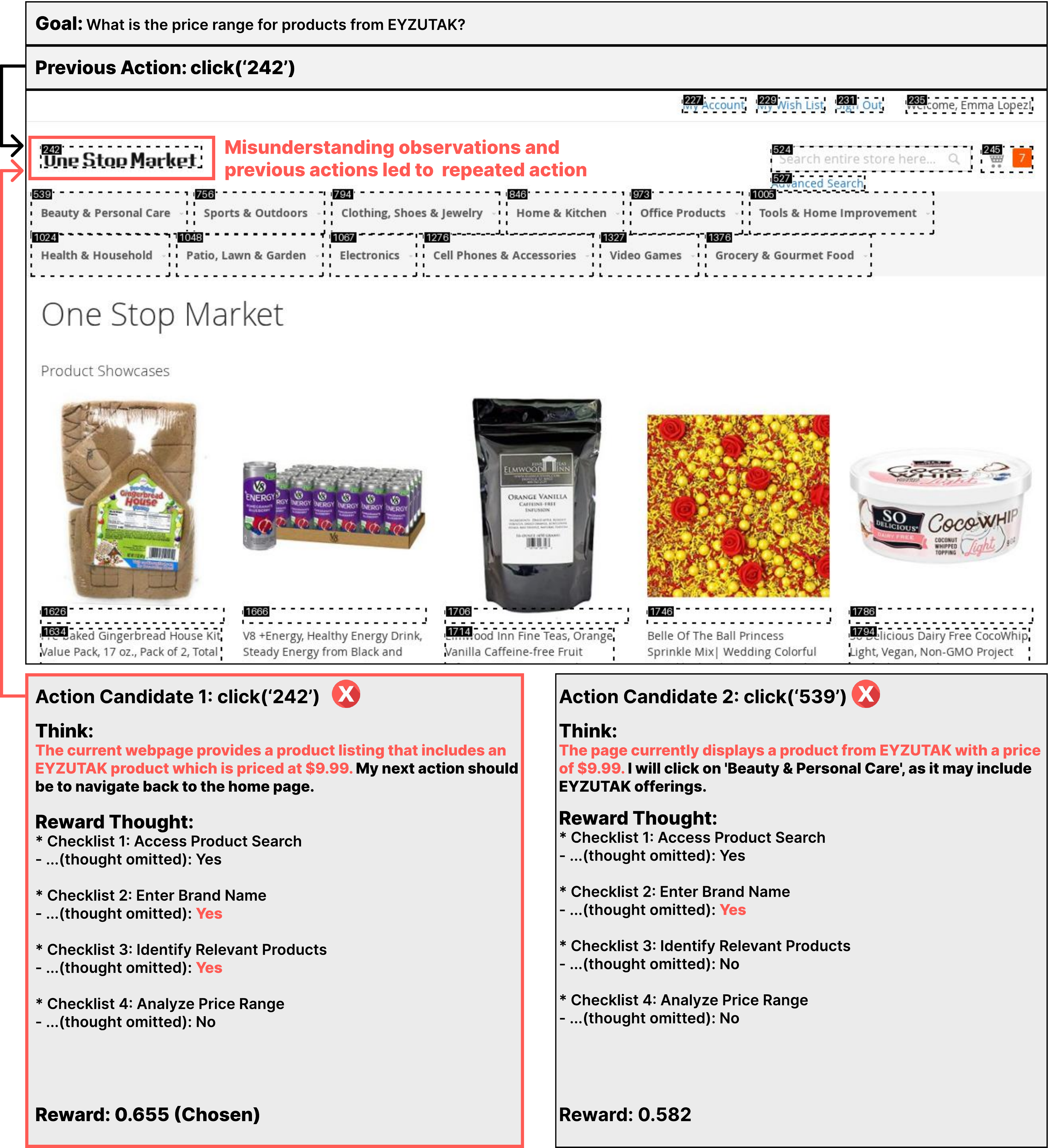}
    \caption{Erroneous Example. By misinterpreting the current observation and ignoring previous actions, the agent performs repetitive actions.}
    \label{fig:case_study_2}
\end{figure}
\begin{figure}[!h]
    \centering
    \includegraphics[width=1.0\linewidth]{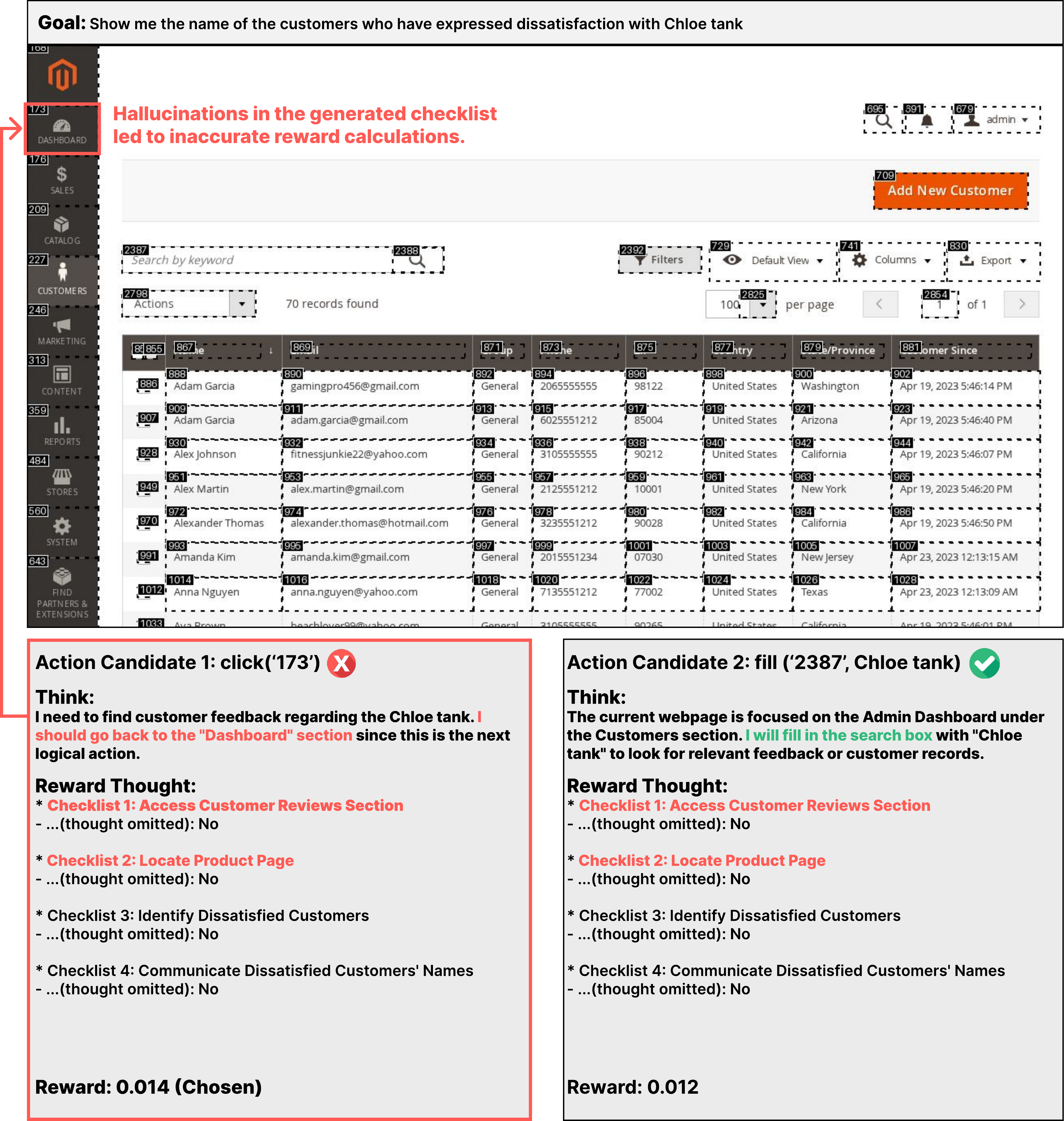}
    \caption{Erroneous Example. When generating the checklist, hallucinations such as navigating to non-existent sections (Customer Reviews Section) or referencing pages that don’t exist (Product Page) lead to incorrect reward calculations.}
    \label{fig:case_study_3}
\end{figure}

\section{Prompts Used in Our Works}
\begin{figure*}[h!]
\centering
\resizebox{0.8\textwidth}{!}{
\begin{tcolorbox}[
colframe=darkgray,        
colback=gray!10,
arc=2mm,
boxrule=1pt,   
title=\textbf{Prompts},
fonttitle=\bfseries
]

You are an expert evaluator of web agent.\\
Your task is to assess how helpful a given agent's THOUGHT and ACTION is in making progress toward the user's goal, based on the current state of the webpage.\\
\\
\# Action space:\\
\texttt{[Description of Action space]}\\
\\
\# Task Description\\
Evaluate how helpful the given thought and action is for achieving the goal.\\
Use the following scale:\\
\\
**Scoring Criteria (1 to 5):**\\
- **5 (Very Helpful)**: The action directly and effectively moves toward fulfilling a key part of the goal.
\\
- **4 (Helpful)**: The action contributes meaningfully to progress, though it may require follow-up actions.
\\
- **3 (Somewhat Helpful)**: The action is partially relevant or a preparatory step, but doesn’t make immediate progress.
\\
- **2 (Slightly Helpful)**: The action is weakly related to the goal or might only indirectly help.
\\
- **1 (Not Helpful)**: The action is unrelated, redundant, or distracts from the goal.
\\
\\
\# Given Information\\
\#\# User Instruction\\
\texttt{\{intent\}}\\
\\
\#\# Trajectory\\
\texttt{\{trajectory\}}\\
\\
\#\# Current State\\
\\
\#\#\# Current URL\\
\texttt{\{current\_url\}}\\
\\
\#\#\# AXTREE \\
Note: \texttt{[bid]} is the unique alpha-numeric identifier at the beginning of lines for each element in the AXTree. Always use bid to refer to elements in your actions. \\
\texttt{\{text\_observation\}} \\

\#\#\# SOM Image Screenshot \\
Here is a current image screenshot of the page, it is annotated with bounding boxes and corresponding bids:\\
<IMAGE\_PLACEHOLDER>\\

\#\# Agent's Response\\
\\
THOUGHT:\\ 
\texttt{\{thought\}}\\
\\
ACTION:\\ 
\texttt{\{action\}}\\
\\
\# Output Format\\
Please return your response in the following format:\\
\\
REASON:\\
\texttt{[Your explanation for the score]}\\
\\
SCORE:\\
\texttt{[1-5]}\\

\end{tcolorbox}
}
\caption{Prompt used to assign reward  with Likert-scale for baseline models.}
\label{fig:prompt_likert}
\end{figure*}

\begin{figure*}[t!]
\centering
\resizebox{0.88\textwidth}{!}{
\begin{tcolorbox}[
colframe=darkgray,        
colback=gray!10,
arc=2mm,
boxrule=1pt,   
title=\textbf{Prompts},
fonttitle=\bfseries
]

You are an AI assistant tasked with generating structured checklists that highlight key subgoals necessary to complete a task.\\
\\
\#\# Task Description\\
\\
User Instruction (Goal): \texttt{\{intent\}}\\
\\
Start Website URL: \texttt{\{start\_url\}}\\
\\
Guidelines for Checklist Generation\\
\\
1. Identify Essential High-Level Subgoals:\\
- A subgoal should represent a significant step involving user interaction that leads to \\noticeable page transitions or meaningful changes in system state.\\
\\
- Consolidate closely related user actions (such as applying multiple filters or selecting several options) into a single subgoal, rather than separate checklist items for each action.\\
\\
- Prioritize only the most critical interactions necessary for meaningful progression, avoiding the inclusion of minor or unnecessary steps (e.g., scroll, hover).\\
\\
\\
2. Provide a Concise Subgoal Analysis: \\
- Before creating the checklist, offer a brief paragraph summarizing the main subgoals, emphasizing significant transitions or page-level interactions.\\
\\
\\
3. Ensure Clear Goal:\\
- If multiple related interactions occur (e.g., setting filters 1, 2, and 3), combine them into one subgoal with clear criteria verifying all required conditions.\\
\\
- The checklist should contain only essential steps, explicitly excluding unnecessary actions, and should not exceed five critical subgoals. It is not necessary to use all five checklist items if fewer steps adequately represent the essential subgoals.\\
\\
\\
\#\#\# Output Format\\
\\
Before generating the checklist, first produce a concise subgoal analysis in a single paragraph summarizing the required interactions. Then, based on this, generate the checklist following the format below:\\
\\
\texttt{[SUBGOAL ANALYSIS]} \\
\texttt{[One-paragraph summary explaining the key subgoals and their logical sequence in task completion.]} \\
\\
\texttt{[CHECKLISTS]}\\
Checklist X:\\
\texttt{[Short title of the action/goal]}\\
\\
- Goal: \\
\texttt{[Brief description of the subgoal at this stage, emphasizing the purpose of the action.]}

\end{tcolorbox}
}
\caption{Prompt used to generate checklist for baseline models.}
\label{fig:prompt_checklist_generation}
\end{figure*}

\begin{figure*}[t!]
\centering
\resizebox{0.88\textwidth}{!}{
\begin{tcolorbox}[
colframe=darkgray,        
colback=gray!10,
arc=2mm,
boxrule=1pt,   
title=\textbf{Prompts},
fonttitle=\bfseries
]

You are an expert evaluator of web agent.\\
Your task is to assess how helpful a given agent's THOUGHT and ACTION is in making progress toward the user's goal, based on the current state of the webpage.\\
\\
\# Action space:\\
\texttt{[Description of Action space]}\\
\\
\# Task Description\\
Your task is to evaluate how well the agent's THOUGHT and ACTION satisfy each item in the checklist.\\
Use the task instruction, trajectory (including previously completed steps from history), current webpage state, and the agent's current response as evidence for your evaluation.\\
For each checklist item:\\
- Mark it as `Yes' if it is clearly and fully satisfied either in the current response or already completed in the history.\\
- Mark it as `In Progress' if the agent has made partial but meaningful progress toward completing the item.\\
- Mark it as `No' if there is ambiguity, insufficient evidence, or the step is incomplete or not yet started.\\
\\
\# Given Information\\
\#\# User Instruction\\
\texttt{\{intent\}}\\
\\
\#\# Trajectory\\
\texttt{\{trajectory\}}\\
\\
\#\# Current State\\
\#\#\# Current URL\\
\texttt{\{current\_url\}}\\
\\
\#\#\# AXTREE\\
Note: \texttt{[bid]} is the unique alpha-numeric identifier at the beginning of lines for each element in the AXTree. Always use bid to refer to elements in your actions.\\
\texttt{\{text\_observation\}}
\\
\#\#\# SOM Image Screenshot\\
Here is a current image screenshot of the page, it is annotated with bounding boxes and corresponding bids:\\
<IMAGE\_PLACEHOLDER>\\
\\
\#\# Agent's Response\\
THOUGHT:\\ 
\texttt{\{thought\}}\\
ACTION:\\
\texttt{\{action\}}\\
\\
\#\# Output Format\\
Please return your response in the following format:\\
\\
REASON:\\ 
\texttt{[Write a single, coherent paragraph explaining how well the agent's response satisfies the checklist overall. Use both the history and the agent's current thought/action as evidence. Mention specific strengths or missing elements that influence your decision.]}\\
\\
CHECKLIST EVALUATION:\\
Checklist X:\\ 
\texttt{[Yes / In Progress / No]}
\end{tcolorbox}
}
\caption{Prompt used to assign rewards with checklist for baseline models.}
\label{fig:prompt_checklist_reward}
\end{figure*}

\begin{figure*}[t!]
\centering
\resizebox{0.88\textwidth}{!}{
\begin{tcolorbox}[
colframe=darkgray,        
colback=gray!10,
arc=2mm,
boxrule=1pt,   
title=\textbf{Prompts},
fonttitle=\bfseries
]
You are an AI assistant tasked with generating structured checklists that highlight key subgoals necessary to complete a task.\\
\\
\# Task Description\\
Generate a checklist which are key milestones for achieving the given instruction.\\ 
Frist, provide a concise subgoal analysis in a single paragraph summarizing the required interactions.\\ 
Then, based on this, generate the checklist with brief description.\\
\\
\# Given Information\\
\#\# User Instruction\\
\texttt{\{intent\}}\\
\\
\#\# Current URL\\
\texttt{\{start\_url\}}
\end{tcolorbox}
}
\caption{Prompt used to generate checklist for \method.}
\label{fig:prompt_checklist_generation_ours}
\end{figure*}

\begin{figure*}[t!]
\centering
\resizebox{0.88\textwidth}{!}{
\begin{tcolorbox}[
colframe=darkgray,        
colback=gray!10,
arc=2mm,
boxrule=1pt,   
title=\textbf{Prompts},
fonttitle=\bfseries
]
You are an expert evaluator of web agent.\\
Your task is to assess how helpful a given agent's THOUGHT and ACTION is in making progress toward the user's goal, based on the current state of the webpage.\\
\\
\# Task Description\\
Evaluate how well the agent’s THOUGHT and ACTION satisfy each item in the checklist using the task instruction, trajectory (including previously completed steps), current webpage state, and the agent’s latest response.\\
\\
Start by writing a concise paragraph summarizing the agent’s overall performance.\\
\\
Refer to the reasoning provided in the trajectory, and discuss whether the THOUGHT is appropriate and the ACTION moves the task forward.\\
\\
Then, assess each checklist item individually using the following labels:\\
- Yes: The item is fully and clearly satisfied, either in the current response or previously completed.\\
- In Progress: There is meaningful partial progress toward completing the item.\\
- No: The item is not satisfied due to ambiguity, insufficient evidence, or lack of progress.\\
\\
\# Given Information\\
\#\# User Instruction\\
\texttt{\{intent\}}\\
\\
\#\# Trajectory\\
\texttt{\{trajectory\}}\\
\\
\#\# Current State\\
\#\#\# Current URL\\
\texttt{\{current\_url\}}\\
\\
\#\#\# AXTREE\\
Note: \texttt{[bid]} is the unique alpha-numeric identifier at the beginning of lines for each element in the AXTree.Always use bid to refer to elements in your actions.\\
\texttt{\{text\_observation\}}\\
\\
\#\#\# SOM Image Screenshot\\
Here is a current image screenshot of the page, it is annotated with bounding boxes and corresponding bids:\\
<IMAGE\_PLACEHOLDER>\\
\\
\#\# Checklist\\
\texttt{\{checklist\}}\\
\\
\#\# Agent's Response\\
THOUGHT:\\
\texttt{\{thought\}}\\
\\
ACTION:\\
\texttt{\{action\}}
\end{tcolorbox}
}
\caption{Prompt used to assign rewards for \method.}
\label{fig:prompt_reward_ours}
\end{figure*}

\begin{figure*}[h!]
\centering
\resizebox{0.9\textwidth}{!}{
\begin{tcolorbox}[
colframe=darkgray,        
colback=gray!10,
arc=2mm,
boxrule=1pt,   
title=\textbf{Prompts},
fonttitle=\bfseries,
fontupper=\footnotesize
]
\# Instructions\\
You are given a draft thought and action for the current step. This draft has not been executed yet.\\
It was evaluated by a reward model using a checklist based on the task goals.\\
Your task is to reflect on the checklist-based feedback and improve the proposed action.\\
Based on the page state and the provided feedback, revise the thought if needed and produce a better action that will be executed.\\
Your answer will be interpreted and executed by a program, so be precise and follow the formatting instructions.\\
\\
\# Goal:\\
\texttt{\{intent\}}\\
\\
\# Current State\\
\#\# Current URL:\\
\texttt{\{current\_url\}}\\
\\
\#\# AXTree:\\
Note: [bid] is the unique alpha-numeric identifier at the beginning of lines for each element in the AXTree. Always use bid to refer to elements in your actions.\\
Note: only elements that are visible in the viewport are presented. You might need to scroll the page, or open tabs or menus to see more.\\
Note: You can only interact with visible elements. If the "visible" tag is not present, the element is not visible on the page.\\
\texttt{\{text\_observation\}}\\
\\
\# History of interaction with the task:\\
\texttt{\{trajectory\}}\\
\\
\# Action space:\\
\texttt{[Description of Action space]}\\
\\
\# Draft Thought and Action:\\
Thought: \texttt{\{thought\}}\\
Action: \texttt{\{action\}}\\
\\
\# Reward Model Feedback:\\
The reward model evaluates actions using a checklist derived from task-specific goals. Each checklist item represents a key subgoal or intermediate step.\\
\\
Feedback:\\
<feedback>\\
\texttt{\{feedback\}}\\
</feedback>\\
\\
\# Concrete Examples\\
Here is a concrete example of how to format your answer.\\
Make sure to follow the template with proper tags:\\
\\
<EXAMPLE\_PLACEHOLDER>\\
\# Your Response:\\
\texttt{
<think>\\
To move toward Checklist 1—accessing the **Showerthoughts** forum—I should first make it easier to locate that forum in the long list. Clicking the **“Alphabetical”** link will reorder all forums alphabetically, so I can then quickly scroll to “Showerthoughts” and open it. This directly progresses us to the first checklist item.\\
</think>\\
<action>\\
click('117')\\
</action>\\
}
\end{tcolorbox}
}
\caption{Prompt used to generate a refined action for Refinement.}
\label{fig:prompt_reflexion}
\end{figure*}

\begin{figure*}[t!]
\centering
\resizebox{0.88\textwidth}{!}{
\begin{tcolorbox}[
colframe=darkgray,        
colback=gray!10,
arc=2mm,
boxrule=1pt,   
title=\textbf{Prompts},
fonttitle=\bfseries
]
You are an expert evaluator of web agent. Your task is to assess how helpful a given agent's THOUGHT and ACTION is in making progress toward the user's goal, based on the current state of the webpage.\\
\\
\# Action space:\\
\texttt{[Description of Action space]}\\
\\
\# Given Information\\
\#\# User Instruction\\
\texttt{\{intent\}}\\
\\
\#\# Trajectory\\
\texttt{\{trajectory\}}\\
\\
\#\# Current State\\
\#\#\# Current URL\\
\texttt{\{current\_url\}}\\
\\
\#\#\# AXTREE\\
Note: [bid] is the unique alpha-numeric identifier at the beginning of lines for each element in the AXTree. Always use bid to refer to elements in your actions.\\
\texttt{\{text\_observation\}}\\
\\
\#\# Agent's Response\\
THOUGHT: \texttt{\{thought\}}\\
ACTION: \texttt{\{action\}}\\
\\
\# Output Format:\\
\texttt{
Please return your response in the following format:\\
REASON: [Your explanation for the score]\\
SCORE: [1-5]
}

\end{tcolorbox}
}
\caption{Prompt used to assign rewards with Likert-scale for baseline models in trajectory search.}
\label{fig:prompt_eval_likert_bon}
\end{figure*}

\begin{figure*}[t!]
\centering
\resizebox{0.88\textwidth}{!}{
\begin{tcolorbox}[
colframe=darkgray,        
colback=gray!10,
arc=2mm,
boxrule=1pt,   
title=\textbf{Prompts},
fonttitle=\bfseries
]
You are an expert evaluator of web agent. Your task is to assess how helpful a given agent's THOUGHT and ACTION is in making progress toward the user's goal, based on the current state of the webpage.\\
\\
\# Action space:\\
\texttt{[Action Space Description]}

\# Task Description\\
Your task is to evaluate how well the agent's THOUGHT and ACTION satisfy each item in the checklist.\\
Use the task instruction, trajectory (including previously completed steps from history), current webpage state, and the agent's current response as evidence for your evaluation. Clearly consider any items already successfully completed or currently in progress according to the provided trajectory.\\
For each checklist item:\\
- Mark it as `Yes' if it is clearly and fully satisfied either in the current response or already completed in the history.\\
- Mark it as `In Progress' if the agent has made partial but meaningful progress toward completing the item.\\
- Mark it as `No' if there is ambiguity, insufficient evidence, or the step is incomplete or not yet started.\\
\\
\# Given Information\\
\#\# User Instruction\\
\texttt{\{intent\}}\\
\\
\#\# Trajectory\\
\texttt{\{trajectory\}}\\
\\
\#\# Current State\\
\#\#\# Current URL\\
\texttt{\{current\_url\}}\\
\\
\#\#\# AXTREE\\
Note: [bid] is the unique alpha-numeric identifier at the beginning of lines for each element in the AXTree. Always use bid to refer to elements in your actions.\\
\texttt{\{text\_observation\}}\\
\\
\#\# Checklist\\
\texttt{\{checklist\}}\\
\\
\#\# Agent's Response\\
THOUGHT: \texttt{\{thought\}}\\
ACTION: \texttt{\{action\}}\\
\\
\# Output Format\\
Please return your response in the following format:\\
\texttt{
REASON: [Write a single, coherent paragraph explaining how well the agent's response satisfies the checklist overall. Use both the history and the agent's current thought/action as evidence. Mention specific strengths or missing elements that influence your decision.]\\
CHECKLIST EVALUATION:\\
Checklist X: [Yes / In Progress / No]\\
}
\end{tcolorbox}
}
\caption{Prompt used to assign rewards with checklists for baseline models in trajectory search.}
\label{fig:prompt_eval_checklist_bon}
\end{figure*}

\begin{figure*}[t!]
\centering
\resizebox{0.88\textwidth}{!}{
\begin{tcolorbox}[
colframe=darkgray,        
colback=gray!10,
arc=2mm,
boxrule=1pt,   
title=\textbf{Prompts},
fonttitle=\bfseries
]

\#\# Instruction \\
You are an expert evaluator tasked with assessing checklists for goal-directed web navigation.\\ 
These checklists are designed to guide an agent through multi-step tasks on a website.\\  
Each checklist consists of subgoals, presented step-by-step, with brief descriptions explaining the purpose of each step. Using the provided intent, start URL, and the reference checklist, evaluate the quality of the checklist according to the following criterion.\\
\\
Criteria: Checklist Validity\\
- Does the checklist contain only valid, relevant, and logically consistent steps that align with the intent and the reference checklist, without introducing incorrect or misleading actions?\\
\\
Using the rubric below, provide a brief justification and assign a score from 1 to 5 (number only, where 1 = very poor and 5 = excellent).\\
\\
Rubric:\\
-1: Very poor: Checklist contains multiple invalid, irrelevant, or misleading steps that conflict with the intent or contradict the reference checklist.\\
\\
-2: Poor: Checklist includes some valid steps but also contains serious logical errors or clearly irrelevant actions that compromise task validity.\\
\\
-3: Fair: Most steps are reasonable and aligned with the task, but there are one or two questionable or weakly justified steps that reduce overall reliability.\\
\\
-4: Good: Checklist is mostly valid and logically sound, with only minor issues such as slight ambiguities or borderline-relevant steps.\\
\\
-5: Excellent: All steps are valid, relevant, and logically consistent with the intent and reference checklist, with no incorrect or misleading content.\\
\\
Respond in the following format:\\
Justification:\\
Score:\\
\\
\#\# Input\\
Intent:\\
\texttt{\{intent\}}\\
\\
start\_url:\\
\texttt{\{start\_url\}}\\
\\
reference checklist:\\
\texttt{\{reference\_checklist\}}\\
\\
generated checklist:\\
\texttt{\{generated\_checklist\}}\\
\\
\#\# Output\\

\end{tcolorbox}
}
\caption{Prompt used to evaluate the quality of the generated checklist based on checklist validity.}
\label{fig:prompt_geval_checklist_validity}
\end{figure*}
\begin{figure*}[t!]
\centering
\resizebox{0.88\textwidth}{!}{
\begin{tcolorbox}[
colframe=darkgray,        
colback=gray!10,
arc=2mm,
boxrule=1pt,   
title=\textbf{Prompts},
fonttitle=\bfseries
]

\#\# Instruction \\
You are an expert evaluator tasked with assessing checklists for goal-directed web navigation.\\ 
These checklists are designed to guide an agent through multi-step tasks on a website.\\  
Each checklist consists of subgoals, presented step-by-step, with brief descriptions explaining the purpose of each step. Using the provided intent, start URL, and the reference checklist, evaluate the quality of the checklist according to the following criterion.\\
\\
Criteria: Subgoal Granularity\\
- Are the checklist steps appropriately scoped, neither too fine-grained nor too coarse, and aligned with the level of detail found in the reference checklist?\\
\\
Using the rubric below, provide a brief justification and assign a score from 1 to 5 (number only, where 1 = very poor and 5 = excellent).\\
\\
Rubric:\\
-1: Very poor: Checklist is extremely unbalanced in granularity, with most steps being either overly fine-grained or excessively coarse, making the structure difficult to interpret or use.\\
\\
-2: Poor: There are several steps with inappropriate granularity—too detailed or too broad—and the overall checklist lacks consistency in how actions are broken down.\\
\\
-3: Fair: The checklist has a mix of well-scoped and poorly scoped steps, with a few instances of overly fine or coarse granularity that cause mild disruption in flow.\\
\\
-4: Good: Most steps are appropriately scoped, with only minor inconsistencies in granularity or density that do not significantly hinder readability or execution.\\
\\
-5: Excellent: The generated checklist strikes an appropriate level of granularity—neither too coarse nor too fine—closely resembling the reference checklist. In addition, the progression through the checklist items is relatively uniform in density.\\
\\
Respond in the following format:\\
Justification:\\
Score:\\
\\
\#\# Input\\
Intent:\\
\texttt{\{intent\}}\\
\\
start\_url:\\
\texttt{\{start\_url\}}\\
\\
reference checklist:\\
\texttt{\{reference\_checklist\}}\\
\\
generated checklist:\\
\texttt{\{generated\_checklist\}}\\
\\
\#\# Output\\

\end{tcolorbox}
}
\caption{Prompt used to evaluate the quality of the generated checklist based on subgoal granularity.}
\label{fig:prompt_geval_subgoal_granularity}
\end{figure*}
\begin{figure*}[t!]
\centering
\resizebox{0.88\textwidth}{!}{
\begin{tcolorbox}[
colframe=darkgray,        
colback=gray!10,
arc=2mm,
boxrule=1pt,   
title=\textbf{Prompts},
fonttitle=\bfseries
]

\#\# Instruction \\
You are an expert evaluator tasked with assessing checklists for goal-directed web navigation.\\ 
These checklists are designed to guide an agent through multi-step tasks on a website.\\  
Each checklist consists of subgoals, presented step-by-step, with brief descriptions explaining the purpose of each step. Using the provided intent, start URL, and the reference checklist, evaluate the quality of the checklist according to the following criterion.\\
\\
Criteria: Goal Coverage\\
- Does the checklist comprehensively reflect the key steps necessary to achieve the final goal, as captured in the reference checklist?\\
\\
Using the rubric below, provide a brief justification and assign a score from 1 to 5 (number only, where 1 = very poor and 5 = excellent).\\
\\
Rubric:\\
-1: Very poor: Checklist omits most key steps found in the reference checklist and contains vague, irrelevant, or misleading content.\\
\\
-2: Poor: Checklist includes a few relevant steps, but misses many essential ones from the reference checklist, resulting in a structure that does not support goal completion.\\
\\
-3: Fair: Checklist reflects most major steps from the reference checklist but misses one or two key actions or includes loosely related steps.\\
\\
-4: Good: Checklist includes nearly all essential steps from the reference checklist, with only minor omissions or slight ambiguities in an otherwise coherent structure.\\
\\
-5: Excellent: Checklist fully captures all key steps covered in the reference checklist, with clear subgoals that directly support achieving the final goal.\\
\\
Respond in the following format:\\
Justification:\\
Score:\\
\\
\#\# Input\\
Intent:\\
\texttt{\{intent\}}\\
\\
start\_url:\\
\texttt{\{start\_url\}}\\
\\
reference checklist:\\
\texttt{\{reference\_checklist\}}\\
\\
generated checklist:\\
\texttt{\{generated\_checklist\}}\\
\\
\#\# Output\\

\end{tcolorbox}
}
\caption{Prompt used to evaluate the quality of the generated checklist based on goal coverage.}
\label{fig:prompt_geval_goal_coverage}
\end{figure*}
\newpage

\newpage
\section*{NeurIPS Paper Checklist}

\begin{enumerate}

\item {\bf Claims}
    \item[] Question: Do the main claims made in the abstract and introduction accurately reflect the paper's contributions and scope?
    \item[] Answer: \answerYes{}
    \item[] Justification: The abstract and the introduction well reflect our contributions. Especially, the third and the last paragraph covers our scope and contributions, respectively.

\item {\bf Limitations}
    \item[] Question: Does the paper discuss the limitations of the work performed by the authors?
    \item[] Answer: \answerYes{}
    \item[] Justification: We discuss the limitation of our work in Appendix~\ref{app:limitation}.

\item {\bf Theory Assumptions and Proofs}
    \item[] Question: For each theoretical result, does the paper provide the full set of assumptions and a complete (and correct) proof?
    \item[] Answer: \answerNA{}
    \item[] Justification: We do not have any theoretical results.

    \item {\bf Experimental Result Reproducibility}
    \item[] Question: Does the paper fully disclose all the information needed to reproduce the main experimental results of the paper to the extent that it affects the main claims and/or conclusions of the paper (regardless of whether the code and data are provided or not)?
    \item[] Answer: \answerYes{}
    \item[] Justification: We fully provide the experimental details in Appendix~\ref{app:webprmbench}. In addition, we release the code and data to allow easy reproduction of our results.

\item {\bf Open access to data and code}
    \item[] Question: Does the paper provide open access to the data and code, with sufficient instructions to faithfully reproduce the main experimental results, as described in supplemental material?
    \item[] Answer: \answerYes{} 
    \item[] Justification: The code for our experiments and example data from \dataset and \benchmark are included in the supplemental material. 

\item {\bf Experimental Setting/Details}
    \item[] Question: Does the paper specify all the training and test details (e.g., data splits, hyperparameters, how they were chosen, type of optimizer, etc.) necessary to understand the results?
    \item[] Answer: \answerYes{}
    \item[] Justification: We provide the training and inference hyperparameters in Appendix~\ref{app:webshepherd}.

\item {\bf Experiment Statistical Significance}
    \item[] Question: Does the paper report error bars suitably and correctly defined or other appropriate information about the statistical significance of the experiments?
    \item[] Answer: \answerYes{}
    \item[] Justification: In the experiment in Section~\ref{ssec:BoN} we sample 20 outputs from the policy to allow reliable experiment. 

\item {\bf Experiments Compute Resources}
    \item[] Question: For each experiment, does the paper provide sufficient information on the computer resources (type of compute workers, memory, time of execution) needed to reproduce the experiments?
    \item[] Answer: \answerYes{}
    \item[] Justification: The compute resources used for running the experiments are described in Appendix~\ref{app:webshepherd}.
    
\item {\bf Code Of Ethics}
    \item[] Question: Does the research conducted in the paper conform, in every respect, with the NeurIPS Code of Ethics \url{https://neurips.cc/public/EthicsGuidelines}?
    \item[] Answer: \answerYes{}
    \item[] Justification: We have carefully read the code of ethics.

\item {\bf Broader Impacts}
    \item[] Question: Does the paper discuss both potential positive societal impacts and negative societal impacts of the work performed?
    \item[] Answer: \answerYes{}
    \item[] Justification: We discuss them in Appendix~\ref{app:societal_impacts}.
    
\item {\bf Safeguards}
    \item[] Question: Does the paper describe safeguards that have been put in place for responsible release of data or models that have a high risk for misuse (e.g., pretrained language models, image generators, or scraped datasets)?
    \item[] Answer: \answerYes{}
    \item[] Justification: We describe the verification process for constructing our dataset in Appendix~\ref{app:wprm_collection}.

\item {\bf Licenses for existing assets}
    \item[] Question: Are the creators or original owners of assets (e.g., code, data, models), used in the paper, properly credited and are the license and terms of use explicitly mentioned and properly respected?
    \item[] Answer: \answerYes{}
    \item[] Justification: We include the citations and the explanations for the models, benchmarks, and code used in our work.

\item {\bf New Assets}
    \item[] Question: Are new assets introduced in the paper well documented and is the documentation provided alongside the assets?
    \item[] Answer: \answerYes{}
    \item[] Justification: We provide the details of \dataset and \method in  Appendix~\ref{app:wprm_collection} and Appendix~\ref{app:webshepherd}, respectively.

\item {\bf Crowdsourcing and Research with Human Subjects}
    \item[] Question: For crowdsourcing experiments and research with human subjects, does the paper include the full text of instructions given to participants and screenshots, if applicable, as well as details about compensation (if any)? 
    \item[] Answer: \answerYes{}
    \item[] Justification: We detail the human annotation process in Appendix~\ref{appendix:human_annotation_details}.

\item {\bf Institutional Review Board (IRB) Approvals or Equivalent for Research with Human Subjects}
    \item[] Question: Does the paper describe potential risks incurred by study participants, whether such risks were disclosed to the subjects, and whether Institutional Review Board (IRB) approvals (or an equivalent approval/review based on the requirements of your country or institution) were obtained?
    \item[] Answer: \answerNA{}
    \item[] Justification: All annotation is made by the authors.

\end{enumerate}

\end{document}